\PassOptionsToPackage{dvipsnames, table}{xcolor}
\documentclass[acmlarge]{acmart}

\AtBeginDocument{%
  }

\setcopyright{acmlicensed}
\copyrightyear{2025}
\acmYear{2025}
\acmDOI{XXXXXXX.XXXXXXX}

\acmJournal{TOSEM}
\acmVolume{00}
\acmNumber{00}
\acmArticle{000}
\acmMonth{0}

\usepackage{soul}
\usepackage{booktabs}

\usepackage[most]{tcolorbox}
\tcbuselibrary{theorems}
\usepackage{fontawesome}
\usepackage[dvipsnames]{xcolor}
\definecolor{mypink1}{rgb}{0.858, 0.188, 0.478}
\definecolor{mypink2}{RGB}{219, 48, 122}
\definecolor{mypink3}{cmyk}{0, 0.7808, 0.4429, 0.1412}
\definecolor{mygray}{gray}{0.6}
\usepackage{booktabs}
\usepackage{color}
\usepackage{tabularray}
\usepackage{graphicx}
\usepackage{subcaption}
\usepackage{url}
\definecolor{Silver}{rgb}{0.752,0.752,0.752}
\definecolor{MineShaft}{rgb}{0.121,0.121,0.121}
\usepackage{verbatim} 
\usepackage[normalem]{ulem} 

\begin{document}


\title{How Effective are Generative Large Language Models in Performing Requirements Classification?}








\author{Waad Alhoshan}
\affiliation{%
  \institution{Imam Mohammad Ibn Saud Islamic University (IMSIU)}
  \city{Riyadh}
  \country{Saudi Arabia}}
\email{wmaboud@imamu.edu.sa}

\author{Alessio Ferrari}
\affiliation{%
  \institution{University College Dublin (UCD)}
  \city{Dublin}
  \country{Ireland}}
\email{alessio.ferrari@ucd.ie}

\author{Liping Zhao}
\affiliation{%
  \institution{University of Manchester}
  \city{Manchester}
  \country{United Kingdom}}
\email{liping.zhao@manchester.ac.uk}

\renewcommand{\shortauthors}{Alhoshan et al.}

\begin{abstract}
\textbf{Context:} In recent years, transformer-based large language models (LLMs) have revolutionised natural language processing (NLP), with generative models opening new possibilities for tasks that require context-aware text generation. Requirements engineering (RE) has also seen a surge in the experimentation of LLMs for different tasks, including trace-link detection, regulatory compliance, and others.

\textbf{Problem:} Requirements classification is a common task in RE. While non-generative LLMs like BERT have been successfully applied to this task, there has been limited exploration of generative LLMs. This gap raises an important question: how well can generative LLMs, which produce context-aware outputs, perform in requirements classification?

\textbf{Solution:} In this study, we explore the effectiveness of three generative LLMs—Bloom, Gemma, and Llama—in performing both binary and multi-class requirements classification. We design an extensive experimental study involving over 400 experiments across three widely used datasets (PROMISE NFR, Functional-Quality, and SecReq). Our study focuses on the impact of four core components: models, datasets, prompts, and tasks. Additionally, we compare the best-performing generative LLMs from this study with the top non-generative models from previous works to identify which model type achieves superior performance in requirements classification.

\textbf{Results:} Our findings show that model selection significantly influences performance, with Bloom excelling in precision, achieving a maximum weighted precision score of approximately $0.77$. Gemma excelled in binary classification, particularly in recall, with an average weighted recall rate of $0.51$. Llama offered balanced performance across both precision and recall, making it well suited for both binary and multi-class classification tasks. Prompt formulation plays an essential role, with assertion-based prompts being the most effective. We also observe that generative models demonstrate robustness to dataset variations, maintaining consistent performance despite changes in input text. In comparison to non-generative models (e.g., All-Mini and SBERT), generative models generally performed less effectively in multi-class classification, while non-generative models, particularly All-Mini, outperformed them in binary classification tasks.

\textbf{Contributions:} This paper presents the first comprehensive experimental study of generative LLMs for requirements classification, offering key insights into their strengths and limitations. We highlight the critical role of four core components—models, datasets, prompts, and tasks—on classification performance. Our study concludes that while factors like prompt design and LLM architecture are universally important, others-—such as dataset variations—-have a more situational impact, depending on the complexity of the classification task. This insight can guide future model development and deployment strategies, focusing on optimising prompt structures and aligning model architectures with task-specific needs for improved performance. 

\end{abstract}

\keywords{Large Language Models, LLMs, Generative LLMs, Non-Generative LLMs, Requirements Classification, Requirements Engineering, Experimental Study}


\maketitle

\section{Introduction}
\label{sec:intro}

In recent years, the \textbf{transformer architecture} \cite{Vaswani2017AttentionIA} has revolutionised NLP and reshaped the landscape of the field. Known for its unparalleled efficiency in processing and generating text, the transformer has become the foundational architecture behind the majority of state-of-the-art \textbf{LLMs}, including BERT \cite{devlin2018bert}, ChatGPT \cite{achiam2023gpt}, Llama \cite{touvron2023llama}, Bloom \cite{le2023bloom}, and GPT-3 \cite{dale2021gpt}, among others. At the heart of its power is the \textbf{self-attention} mechanism, which enables each word or token in an input sequence to interact with every other word, capturing complex contextual relationships irrespective of their distance in the sequence. This attention-driven connectivity empowers transformer-based models to tackle a wide array of real-world applications, ranging from chatbots and virtual assistants to content creation, language translation, text summarization, sentiment analysis, education, and medical diagnostics \cite{alto2024building, raiaan2024review}.  
With the rapid advancement of LLMs, their capabilities continue to grow at an unprecedented pace.

To fully harness the potential of LLMs, our previous study \cite{alhoshan2023zero} explored the effectiveness of \textbf{BERT-based models} for requirements classification tasks. Specifically, we applied four BERT-based models (SBERT\footnote{https://sbert.net/}, all-MiniLM-L12-v2 (All-Mini)\footnote{https://docs.pinecone.io/models/all-MiniLM-L12-v2}, SObert \cite{tabassum-etal-2020-code}, and BERT4RE \cite{ajagbe2022retraining}) with zero-shot learning approaches to both binary and multi-class classification tasks, achieving F1 scores ranging from $0.66$ to $0.80$. While these non-generative models, such as BERT-related models, have shown promising performance, their reliance on high-dimensional word embeddings to make predictions may limit their ability to fully adapt to varying contexts. Although non-generative LLMs can sometimes be applied to generate text, their primary objective is not oriented toward text generation. Instead, they primarily rely on tasks such as predicting masked tokens or next sentence prediction during pre-training, which reflects their focus on understanding and encoding context rather than generating human-like, contextually relevant text. This potential limitation highlights the need to explore how generative LLMs, which can dynamically produce text tailored to the input context, could enhance performance in requirements classification.

In this paper, we aim to bridge this gap by investigating the performance of generative LLMs in the context of requirements classification. Our study focusses on four core components that are critical for leveraging the power of generative models: \textbf{1) Models, 2) Datasets, 3) Prompts,} and \textbf{4) Tasks}. Each of these components plays a central role in influencing classification outcomes, and understanding their impact is essential for improving the practical use of generative LLMs. To the best of our knowledge, this is the first comprehensive experimental study of using generative LLMs for requirements classification tasks. Our contributions are highlighted as follows:

First, we examine three state-of-the-art \textbf{generative LLMs} (Llama \cite{touvron2023llama}, Bloom \cite{le2023bloom}, and Gemma \cite{team2024gemma}). These models represent the next step in language-generation capabilities. We explore their unique architectures and how these features differentiate them from non-generative models, and we investigate how these architectures could influence their effectiveness in requirements classification.

Next, we investigate the \textbf{datasets} used for evaluation. Dataset selection plays an important role in model performance, as the structure, domain, and size of the data can significantly influence the results. In this study, we utilise three well-established datasets—PROMISE NFR~\cite{cleland2007automated,jane_cleland_huang_2007_268542}, Functional-Quality~\cite{dalpiaz2019requirements}, and SecReq~\cite{knauss2011supporting,knausseric20214530183}—which have been widely adopted by researchers in the field of RE. 
Additionally, we explore various variations of these datasets to assess how changes to the requirements text and class labels might impact the performance of requirements classification.

We also examine the role of \textbf{prompts}, the input structures that are fed to generative LLMs. The formulation of prompts is an essential factor influencing model output~\cite{white2023prompt,Ronanki2023RequirementsEU}. Our research investigates how different prompt patterns can optimise generative LLMs for requirements classification, ensuring that the models generate the most relevant and accurate classifications.

Lastly, we analyse the \textbf{tasks} themselves, specifically how different types of requirements classification tasks—binary versus multi-class—interact with generative LLMs. Understanding the impact of task structure on generative models is essential for assessing their potential in practical applications, such as software development and industry use cases.

By focusing on these four core components, this paper aims to provide a comprehensive experimental study of how generative LLMs perform in requirements classification, highlighting how each component influences the models' effectiveness. Our findings offer valuable insights for both researchers and practitioners seeking to implement these advanced models in real-world applications.

Finally, we compare the best-performing generative LLMs found in this experimental study with the best-performing non-generative LLMs from our previous work to determine which model type delivers superior performance in requirements classification. 

The key findings of our study are:

\begin{itemize}
    \item The choice of LLMs significantly affects performance of requirements classification.
    \item Prompt formulation plays a critical role in classification results.
    \item Generative LLMs are robust to dataset variations, maintaining consistent performance despite changes in requirement text and label formatting.
    \item When comparing generative LLMs to non-generative models, we found that no single model consistently outperformed the others in binary classification tasks. However, in multi-class classification, non-generative models emerged as the top performers, while generative LLMs struggled to achieve comparable results.
\end{itemize}

The paper is organised as follows: First, Section \ref{sec:req-class} introduces the concept of requirements classification and provides an overview of existing work on using NLP and machine learning approaches for this task. Next, Section \ref{sec:llms} explores the concepts of LLMs, prompts, and the learning approaches associated with the application of LLMs. Following these foundational sections, Section \ref{sec:approach} describes the learning approaches used in our study. Section \ref{sec:methods} presents our research questions and design. Our findings are reported in Sections \ref{sec:res4RQ1} and \ref{sec:res4RQ2}. We then address potential threats to the validity of our research in Section \ref{sec:valid}. Finally, Section \ref{sec:concl} concludes the paper by summarising our contributions.
\section{Requirements Classification}
\label{sec:req-class}

Research on requirements classification using machine learning has been evolving for nearly two decades, with its origins dating back to the mid-2000s~\cite{ZhaoRC2025,zhao2021natural}. This section begins by introducing the concept of requirements classification, followed by a brief overview of the research in the field. We outline the transition from early studies that utilised traditional machine learning techniques to more recent approaches leveraging deep learning and pretrained language models. In this context, we also present our own research contribution, which is detailed in this paper. For comprehensive guidance and practical insights on implementing machine learning techniques for requirements classification, we direct the reader to the work of Zhao and Alhoshan \cite{ZhaoRC2025}.

\subsection{Concepts and Definitions}

Requirements classification is the task of the assignment of the requirements of a software project to a set of categories according to a given classification scheme. This task can be formulated as a \textbf{text classification} problem \cite{deng2019feature}, as software and system requirements are often written in natural language \cite{zhao2021natural,kassab2014state}. Consequently, we define the requirements classification problem as follows:

\begin{definition}[Requirements Classification] \label{RC}
Given a collection of  $\ N$ requirements $\ R = \{r_1, r_2, ..., r_N\} $ and a set of $\ K$ predefined categories$\ C = \{c_1, c_2, ..., c_k\}$ (e.g., functional, non-functional requirements etc.), the problem of requirements classification is concerned with finding a mapping $\ F$  from the Cartesian product $\ R \times C$  to a set $\{True, False\}$, i.e., $\ F : R  \times C \rightarrow \{True, False\}$. Based on this mapping, for each requirement $\ r_i \in R$  and a category $\ c_j \in C$, if $\ F (r_i, c_j) = True$, then $\ r_i$  belongs to category $\ c_j$, otherwise $\ r_i$  does not belong to $c_j$. 
\end{definition}

As with text classification, requirements classification can be performed using supervised machine learning techniques \cite{sebastiani2002machine,kowsari2019text}, whereby a machine learning model is trained on a set of labelled requirements examples or samples to learn the category labels and their corresponding examples. Once trained, the model can be used to automatically classify new, unseen requirements (i.e., the requirements that are not in the training set). The trained machine learning model is generally called a ``classifier'' or a ``classification model''; in the context of requirements classification, however, the model can be called a ``requirements classifier'' or a ``requirements classification model''. Requirements classification under supervised learning can be formally defined as follows:

\begin{definition}[Supervised Requirements Classification]\label{ML4RC}
Given a training dataset that contains $m$ requirements samples $\ R' = \{r'_1, r'_2, ..., r'_m\} $, where every requirement $\ r'_i \in R'$ is associated with (i.e., pre-assigned) a category label $\ c_j \in C$, requirements classification involves training a \textit{classifier} or \textit{classification model} $F'$ for a given requirements classification task to learn the relationship between each pair of the requirement and its label, such that $F'(r'_i, c_j) = True$. After the training, $F'$ can be used to classify $R$, by automatically assigning each requirement $\ r_i \in R$ with a category $\ c_j \in C$, such that $F'(r_i, c_j) = True$. 
\end{definition}

In general, classification tasks fall into three general types \cite{sokolova2009systematic}: \textit{binary classification}, whereby the input is classified into one, and only one, of two non-overlapping categories $(c_1, c_2)$; \textit{multi-class classification}, whereby the input is classified into one, and only one, of $n$ non-overlapping categories $(c_1, c_2, ... c_n)$, where $n \geq 3$; and \textit{multi-labelled classification}, whereby the input is classified into several of $n$ overlapping categories $c_j$. Multi-labelled classification for requirements is not common, as software development requires each requirement to be allocated to one and only one category. Consequently, requirements classification tasks are typically binary or multi-class: Under binary classification, a requirement is classified into one of two categories, such as the functional or non-functional category; under multi-class classification, a requirement is classified into one of more than two categories, such as security, usability and performance.   

\subsection{Research Progress in Requirements Classification}

\subsubsection{Early Approaches Using Traditional Machine Learning Models}

Classic machine learning models, such as Naive Bayes (NB) and Support Vector Machines (SVM) have been widely researched in requirements classification \cite{binkhonain2019review}. Here, we briefly review some of the studies that have applied such models. 

The earliest study was by Cleland-Huang \textit{et al.}~\cite{cleland2007automated}, who applied a probability function similar to NB to distinguish between functional requirements (FRs) and nonfunctional requirements (NFRs), and identify different NFR classes. A set of indicator terms, i.e., keywords, were created as the features to train the classifier. The study showed that the classifier produced a good recall result (up to $80\%$) but a poor precision (up to $21\%$). The dataset used in this study, called ``the PROMISE NFR dataset''~\cite{jane_cleland_huang_2007_268542}, contains 625 requirements, labelled with 12 classes, made up of one FR class and 11 NFR classes. The requirements in this dataset were collected from 15 software projects and labelled primarily by students. The dataset, which remains to be one of the very few labelled dataset suitable for requirements classification\footnote{There is an extended NFR dataset by Lima et al. \cite{lima2019software}.}, has been widely used in the RE research community to train classification models (e.g., \cite{kurtanovic2017automatically,dalpiaz2019requirements,hey2020norbert,alhoshan2022zero,alhoshan2023zero}).

Another early study, by Casamayor \textit{et al.}~\cite{casamayor2010identification}, proposed an iterative learning method similar to active learning to support requirements classification. In each iteration, an NB model was trained using a small number of the training data from the PROMISE NFR dataset and then tuned manually through the user feedback from a human operator. The study showed a good classification performance on most classes, with a maximum precision of above $80\%$ and a maximum recall of above $70\%$ on most classes.

In addition to NFR classification tasks, Knauss \textit{et al.}~\cite{knauss2011supporting} reported a study on security requirements classification, in which the authors trained an NB classifier to identify security-relevant requirements on three datasets from the software industry (the public SeqReq dataset). They achieved a precision of more than $80\%$ and a recall of more than $90\%$. In another work, Riaz \textit{et al.}~\cite{riaz2014hidden} proposed an approach to extract security-relevant sentences from requirements documents. They used a dataset of 10,963 sentences belonging to six different documents from the healthcare domain. The proposed approach was semi-automatic and based on a K-Nearest Neighbour (KNN) model. The authors achieved a precision of $82\%$, and a recall of $79\%$.

After a period of dormancy, Kurtanović and Maalej~\cite{kurtanovic2017automatically} conducted a study in 2017 that examined the effectiveness of an SVM model in automatically classifying requirements from the PROMISE dataset. A total of $1,000$ features were selected using three n-gram models (with $n=1,2,3$) plus Part-of-Speech (POS) tags. The study achieved $92\%$ precision and $90\%$ recall. This study was reproduced by Dalpiaz \textit{et al.}~\cite{dalpiaz2019requirements}. 
Dalpiaz et al. also re-tagged the PROMISE NFR dataset to distinguish between functional and quality aspects, considering that requirements can include both, as also noticed by existing studies~\cite{eckhardt2016non}.

\subsubsection{Advancements in Research with Deep Learning Models} 

One of the early attempts to apply deep learning-based solutions to requirements classification is the work by Winkler and Vogelsang \cite{winkler2016automatic}. The study aimed to implement a deep learning classifier to discriminate requirements from general information. The classifier was built using a Convolutional Neural Network (CNN) architecture. The authors constructed a training dataset using 89 industrial requirement documents. The word2vec language model~\cite{mikolov2013distributed} was used to create 128-dimensional embeddings as vector representations and max-pooling techniques to aggregate the processed features from the layers. The classifier was able to detect requirement classes with an F1-score of 80\% and precision and recall rates of 73\% and 89\%, respectively. The authors reported some limitations of the proposed model, including the difficulty of interpreting the reasons for inaccuracies of the classifier. 

A similar work was presented by Dekhtyar and Fong \cite{dekhtyar2017re}, who used the PROMISE NFR and SeqReq datasets to train a CNN-based classifier using word2vec embeddings. 
On security requirements, the classifier obtained an overall performance of 91\% as an F1-score with a precision and recall rate of 92\% and 91\%, respectively. On the PROMISE NFR dataset, the model yielded an F1-score of 92\% of 92\% and recall and precision of 93\%. 

Similar to the aforementioned studies, Baker \textit{et al.} \cite{baker2019automatic} used CNNs for requirements classification. The authors aimed to classify requirements into five categories: maintainability, operability, performance, security and usability. The training dataset contained about 1,000 NFRs, extracted from the PROMISE dataset and Quality Attributes NFR dataset from RE'17 Data Challenge\footnote{\url{http://ctp.di.fct.unl.pt/RE2017/pages/submission/data\_papers/}}. The CNN model achieved a precision score ranging between 82\% and 94\% and a recall score ranging between 76\% and 97\%, with an F-score ranging between 82\% and 92\%.

The Recurrent Neural Network (RNN) deep learning architecture and its variants, such as Gated Recurrent Unit (GRU) and Long Short-Term Memory (LSTM), have also been employed for requirements classification due to its ability to model contextual dependency in long input sequences. Kaur and Kaur ~\cite{kaur2022sabdm} employed an LSTM network to classify a dataset of 34 industrial requirement specifications in addition to the PROMISE dataset. Then, a Global Vectors for Word Representation (GloVe) word embedding model~\cite{pennington2014glove} was used to convert word tokens into vector representations for the neural network inputs. 
The classifier yielded an overall F1-score of 96\% with precision and recall scores of 95\% and 96\%, respectively. 

Another study by AlDhafer \textit{et al.} \cite{aldhafer2022end} utilized Bidirectional GRU (BiGRU) to classify FRs and NFRs, and distinguish between NFR sub-classes as well. The classifier model was trained using two different datasets, the PROMISE and EHR datasets, where the latter is a pre-labeled dataset from electronic health records~\cite{slankas2013automated}. The model performed really well in classifying between FR and NFR classes with overall performance rates of 94\% as an F1-score, and 93\% and 95\% as precision and recall. For the fine-grained multi-classification of NFR classes, the BiGRU model obtained an acceptable performance with an F1-score of 78\% in comparison to the notable results of 87\% as an F1-score when the classifier trained only on the top-four NFR classes of the training dataset.

\subsubsection{Pioneering Research with New Learning Models}

The first studies of using LLMs for requirements tasks were reported in two papers \cite{hey2020norbert,sainani2020extracting,Wang2020Deep}. 
These studies applied the fine-tuning approach to the BERT model, involving adjusting the model weights based on a task-specific dataset to improve performance on the target task, e.g., requirements classification.  
Hey \textit{et al.}~\cite{hey2020norbert} proposed a transfer learning approach for requirements classification. The approach, named ``NoRBERT'', employed two pre-trained BERT models (BERT\textsubscript{base} and BERT\textsubscript{large}) as classification models. These models were fine-tuned for both binary and multi-class classification tasks. The PROMISE NFR dataset and its class labels were used to train the BERT models.  
The study showed that both fine-tuned BERT\textsubscript{base} and BERT\textsubscript{large} achieved an F1-score of over 90\% for the FR class and 93\% for NFR classes. In comparison with the then state-of-the-art results for requirements classification tasks obtained from the classic and deep learning models, the study outperformed the SVM model used by Kurtanovi\'c and Maalej~\cite{kurtanovic2017automatically}, the NB model used by Abad \textit{et al.}~\cite{abad2017works}, and the CNN model used by Dekhtyar and Fong \cite{dekhtyar2017re}.  

The paper by Sainani \textit{et al.}~\cite{sainani2020extracting} reported a study that explored different machine learning models for extracting and classifying requirements from a dataset of software engineering contract documents. The dataset contained 5472 requirements, labeled with 14 classes, including Project Delivery, Legal Process, Screening/Onboarding, Vendor Corporate, HR Client Policy, HR Laws, and Personnel Allocation. The study applied NB, Random Forest (RF), Support Vector Machines (SVM), BiLSTM, and BERT\footnote{The authors of this study did not specify whether they used BERT\textsubscript{base} or BERT\textsubscript{large}.}. The study showed that the fine-tuned BERT model outperformed the remaining four models, achieving an F1-score of over 80\% for 9 requirements classes. However, the dataset used in this study is not publicly available.

Chatterjee \textit{et al.}~\cite{chatterjee2021pipeline} reported a study on multi-class classification tasks involving a large training set of 2,122 NFRs. The study employed three BiLSTM-based models and a pre-trained BERT\textsubscript{base} model. The study showed that the fine-tuned BERT outperformed all the other models, confirming the power of BERT for the task. The dataset used in this study is not publicly available. 

While fine-tuned language models can outperform both traditional and deep learning models, fine-tuning requires a model to be trained on thousands or tens of thousands of labelled task-specific samples in order to achieve state-of-the-art performance \cite{devlin2018bert,brown2020language}. To address the lack of training data in RE whilst taking advantage of language models, Alhoshan \textit{et al.}~\cite{alhoshan2022zero} presented the first study in RE that used the ZSL approach for the classification of NFRs. This study used the embedding-based ZSL approach to assess the potential of nine well-known models without fine-tuning or training. These models include standard BERT-based models, a light-weight and distilled version of BERT called all-MiniLM-L12-v2~\cite{alhoshan2022zero}, and a larger model called XLNet. A subset consisting of 67 usability and 66 security requirements was extracted from the PROMISE NFR dataset. The study showed that, without fine-tuning or training, these nine models achieved encouraging results. In particular, the all-MiniLM-L12-v2 model achieved a weighted F score of 82\% (86\% was obtained by the fine-tuned BERT\textsubscript{large} model in Hey \textit{et al.}~\cite{hey2020norbert}). 

Following this successful preliminary study, Alhoshan \textit{et al.}~\cite{alhoshan2023zero} carried out a large-scale study involving two datasets (the original PROMISE NFR \cite{jane_cleland_huang_2007_268542} and a security requirements dataset \cite{knausseric20214530183}) with a total of 1020 requirements. Four language models were selected to perform a variety of binary, multi-class and multi-labelled classification tasks without fine-tuning or training. This study reported different label configuration methods to create different semantic-rich labels for each requirement class. The study showed that the ZSL all-MiniLM-L12-v2 outperformed the traditional SVM models (e.g., Kurtanovi\'c and Maalej~\cite{kurtanovic2017automatically}) and achieved comparable performance to the fine-tuned BERT\textsubscript{large} model in Hey et al.~\cite{hey2020norbert}. 

In addition to ZSL, a recent study also explored few-shot learning (FSL)~\cite{nayak2023few}, an approach that utilises a small number of examples to guide a language model to perform a specific task. This study proposed five FSL approaches to classify software requirements for the 3098 BOSCH automotive domain according to three patterns---i.e., the classes for this case. The training set contained 15 example requirements per pattern, and each FSL approach was repeated with three random seeds (samples) for training data selection. The study showed that with 15 seed samples for training per category, the five approaches achieved an accuracy between 82.66\% and 89.00\%. The dataset used in this study is not publicly available. 

\subsection{Our Contribution}

This paper presents the first comprehensive experimental study of generative LLMs for requirements classification, providing key insights into their strengths and limitations. We focus on the impact of four core components: models, datasets, prompts, and tasks. Our findings highlight that model selection significantly affects performance, prompt formulation is essential for optimising results, and generative models demonstrate robustness to dataset variations, maintaining consistent performance even with changes in input. Additionally, we compare the best-performing generative LLMs from this study with the top non-generative models from our previous work \cite{alhoshan2023zero}, offering a clear comparison of which model type achieves superior performance in requirements classification. This work lays the groundwork for further exploration of generative LLMs in the field of requirements engineering.

\section{Large Language Models}
\label{sec:llms}

This section introduces the key concepts of transformers, generative and non-generative LLMs, prompt engineering, and learning approaches that are integral to their application.

\subsection{Transformer Architecture}

LLMs are based on the \textbf{transformer architecture}~\cite{Vaswani2017AttentionIA}, a novel framework that incorporates the \textbf{attention} mechanism. This mechanism allows the model to effectively capture complex relationships between tokens within an input sequence, even when they are separated by long distances, enabling it to better understand contextual dependencies. Three primary attention mechanisms are used in LLMs:

\begin{itemize}
    \item \textbf{Self-Attention:} First proposed in~\cite{Vaswani2017AttentionIA} where each word in the sequence focusses on every other word, allowing for context-aware understanding. Although highly accurate, it can be computationally expensive, especially for longer texts.

    \item \textbf{Multi-Query Attention:} Intially proposed by~\cite{shazeer2019fast} where words are assigned multiple attention ``spotlights'' to process different aspects of the text in parallel. While faster, it may sacrifice some detail for efficiency.

    \item \textbf{Grouped-Query Attention:} As introduced in ~\cite{ainslie2023gqa} where words are grouped into categories, and each group shares a single spotlight. This method strikes a balance between efficiency and detail, maintaining context while reducing computational load.
\end{itemize}

Transformers serve as the backbone of prominent models like BERT~\cite{devlin2018bert}, GPT~\cite{radford2019language}, and T5~\cite{Raffel2020}, which have become the dominant architecture in modern NLP. These models power some of the most advanced LLMs today, including ChatGPT~\cite{achiam2023gpt}, Llama~\cite{touvron2023llama}, Gemma~\cite{team2024gemma}, and Bloom~\cite{le2023bloom}.

LLMs have demonstrated remarkable success across various domains, from chatbots and virtual assistants to content creation, translation, and medical diagnostics. In recent years, specialised \textit{Code LLMs}, such as StarCoder~\cite{li2023starcoder} and CodeLlama~\cite{roziere2023code}, have gained traction in the software industry for code generation. Tools such as GitHub Copilot\footnote{\url{https://github.com/features/copilot}} leverage generative LLMs to assist developers with code generation, bug detection, and improvement suggestions.

\subsection{Generative vs. Non-Generative LLMs}

Recent LLMs, including GPT, T5, ChatGPT, Llama, Gemma, and Bloom, are \textit{designed} with \textbf{generative capabilities}, meaning they can generate new text based on learned patterns. These \textbf{generative LLMs} produce coherent text—-ranging from sentences to entire documents—-that reflects the language they've been trained on. They are also adept at language understanding tasks like analysis, classification, and prediction based on input text.

In contrast, \textbf{non-generative LLMs} are optimised primarily for language understanding tasks. These models focus on discriminative tasks such as analysis, classification, and prediction, without generating new text. Early models like BERT, RoBERTa \cite{liu2019roberta}, SBERT\footnote{https://sbert.net/}, and all-MiniLM-L12-v2\footnote{https://docs.pinecone.io/models/all-MiniLM-L12-v2} are examples of non-generative LLMs.

While many non-generative LLMs do have the potential to generate text (especially when fine-tuned, e.g.,~\cite{zhang2019bertscore}), their primary design and optimisation revolve around text understanding rather than generation. The key difference lies in whether the model is intended for text generation or optimised for language understanding tasks. Table \ref{tab:generative} summarises the differences between generative and non-generative LLMs.

\begin{table}
\centering
\caption{Comparison of Generative and Non-Generative LLMs}
\label{tab:generative}
\resizebox{\textwidth}{!}{
\begin{tabular}{l|l|l} 
\toprule
\multicolumn{1}{c|}{\textbf{Aspect}} & \multicolumn{1}{c|}{\textbf{Generative Models}}                & \multicolumn{1}{c}{\textbf{Non-Generative Models}}                          \\ 
\hline
\textbf{Main Task}                   & Text generation, creative writing, free-form text              & Text understanding, classification, question answering                      \\
\textbf{Examples}                    & GPT-3, GPT-4, T5, ChatGPT, Llama, Gemma, Bloom                 & BERT, RoBERTa, SBERT, all-MiniLM-L12-v2)                                    \\
\textbf{Core Objective}              & Generate text based on context, prompts, or input              & Understand and analyze input text, classification tasks                     \\
\textbf{Training Method}             & Typically autoregressive (predicting next word)                & Often masked language modeling (MLM) for understanding                      \\
\textbf{Text Generation Capability}  & Primary focus on generating fluent, coherent text              & Can generate text but not optimized for it                                  \\
\textbf{Usage}                       & Dialogue systems, content creation, translation, summarization & Sentiment analysis, question answering, classification, entity recognition  \\
\bottomrule
\end{tabular}
}
\end{table}

\subsection{Prompt Engineering}

A \textbf{``prompt''} is an input description or instruction provided to a LLM to guide its response. Since LLMs generate outputs based on the inputs they receive, the design of a prompt can significantly influence the quality and relevance of the model's output~\cite{mizrahi2024state}. Therefore, crafting clear, specific, and contextually rich prompts is important for effective interaction with LLMs. \textbf{Prompt engineering} refers to the practice of creating prompts that effectively communicate with LLMs to generate desired responses.

To aid in prompt formulation, researchers have designed various \textbf{prompt patterns} \cite{white2023prompt}.  A prompt pattern is a reusable prompt structure or template that can be adapted for different applications. Each prompt pattern is given a distinct name for easy identification and cross-application sharing. For instance, Ronanki et al. \cite{Ronanki2023RequirementsEU} developed five prompt patterns for requirements classification: ``Cognitive Verifier'', ``Context Manager'', ``Persona'', ``Question Refinement'', and ``Template''.  

Prompts can take several forms~\cite{white2023prompt}, including: 

\begin{itemize}

    \item \textbf{Instructional Prompts:} Directly instruct the model to perform a specific task (e.g., ``Summarize this text'').
    \item \textbf{Conversational Prompts:} Mimic dialogue or question-answering to encourage a natural response (e.g., ``What do you think about…?'').
    \item \textbf{Examples or Few-Shot Learning:} Providing examples within the prompt to demonstrate the desired output format.
    \item \textbf{Iterative Refinement:} Effective prompting often involves trial and error. By analysing the responses generated, users can refine their prompts to improve accuracy and relevance.
\end{itemize}

It is important to note that prompting is not merely about instructing or requesting outputs in a ChatGPT-like style. When applied to LLMs that are not fine-tuned for instruction-following or chat-like tasks, prompts may take the form of assertions to be completed (e.g., ``This requirement: \{requirement text\} is about \{candidate label\}''). In Section \ref{sec:methods}, we propose a set of prompt patterns specifically for requirements classification.

\subsection{LLMs and Zero-Shot Learning}
\label{sec:zsl}

Recent advances in LLMs and their success across various NLP tasks have sparked interest in applying these models to the zero-shot learning (ZSL) paradigm. Given that LLMs are trained on vast and diverse datasets, it is reasonable to assume they can perform ZSL tasks with little to no additional training~\cite{kojima2022large}.

In this context, Wei et al.~\cite{wei2022finetuned} introduced an instruction-tuning method that teaches LLMs to perform unseen tasks by providing task descriptions in natural language. Their findings demonstrated that models trained exclusively on instructions could handle a wide range of NLP tasks, including natural language inference, language translation, summarization, commonsense reasoning, and question answering. The model's performance was comparable to that of traditionally supervised learning models.

Similarly, Kojima et al.~\cite{kojima2022large} showed that a simple prompt, ``Let's think step by step,'' could encourage LLMs to reason through various tasks. This approach, known as a zero-shot prompt, elicited a chain of thought from the models, allowing them to generate plausible solutions without additional training. These results highlight that LLMs are not only \textbf{zero-shot learners} but also \textbf{zero-shot reasoners}, capable of performing complex reasoning tasks without task-specific training.

There are two popular ZSL approaches for LLMs: \textbf{embedding-based learning} \cite{wang2019survey} and \textbf{inference-based learning} (also known as \textbf{entailment-based}) \cite{yin-etal-2019-benchmarking}.

\textbf{Embedding-based learning} focusses on learning a projection or embedding function that associates text features from seen classes with corresponding semantic vectors. The learned function is then used to recognise unseen classes by comparing the similarity between prototype representations and predicted representations of the text samples in the embedding space~\cite{Pourpanah2023review}. This approach has been widely applied to traditional machine learning models for classification tasks such as entity recognition~\cite{ma2016label}, relation extraction~\cite{levy2017zero}, document classification~\cite{nam2016all}, and text classification~\cite{pushp2017train}. Our earlier research applied this approach to requirements classification~\cite{alhoshan2022zero}, \cite{alhoshan2023zero}, where we tested the performance of this learning approach with various traditional language models.

In contrast, \textbf{inference-based learning} focusses on understanding and reasoning about relationships between classes. This approach uses pre-trained knowledge of structured relationships—such as class hierarchies or attributes—to predict connections between seen and unseen classes. By leveraging this reasoning ability, models can make inferences about new classes, even without direct exposure to them during training.

However, \textbf{embedding-based learning} is typically applied to \textbf{non-generative LLMs}, which are trained for specific tasks like Masked Language Modeling (MLM) or Next Sentence Prediction (NSP). These models excel at capturing contextual relationships in text, which is useful for tasks like classification.

In contrast, \textbf{generative LLMs} are more suited to \textbf{inference-based learning}, as these models are adept at reasoning and generating contextually relevant outputs. This makes them particularly effective for tasks that require deeper understanding or handling of complex, unseen examples.
These two learning approaches have the following characteristics:

Furthermore, \textbf{embedding-based learning} tends to be simpler in terms of implementation, often leveraging existing embedding techniques and distance metrics. \textbf{Inference-based learning}, however, may require more sophisticated reasoning mechanisms or additional training on relationships and attributes, making it potentially more complex.

\section{Learning Approaches used in Our Research}
\label{sec:approach}

In this paper, we adopt the two aforementioned learning approaches for requirements classification. Specifically, we apply inference-based learning to generative LLMs and embedding-based learning to non-generative LLMs. The following sections provide an overview of the specific methods employed for each of these approaches.

\subsection{Inference-Based Learning}
\label{sec:IBL}

For applying generative LLMs to requirements classification, we utilize an inference-based learning approach. This method involves three key steps: \textbf{model prompting and input encoding}, \textbf{inferencing}, and \textbf{prediction}. These steps are depicted in Figure \ref{fig:zsl-approach} and are described in more detail below:

\begin{figure}
    \centering
    \includegraphics[width=0.9\linewidth]{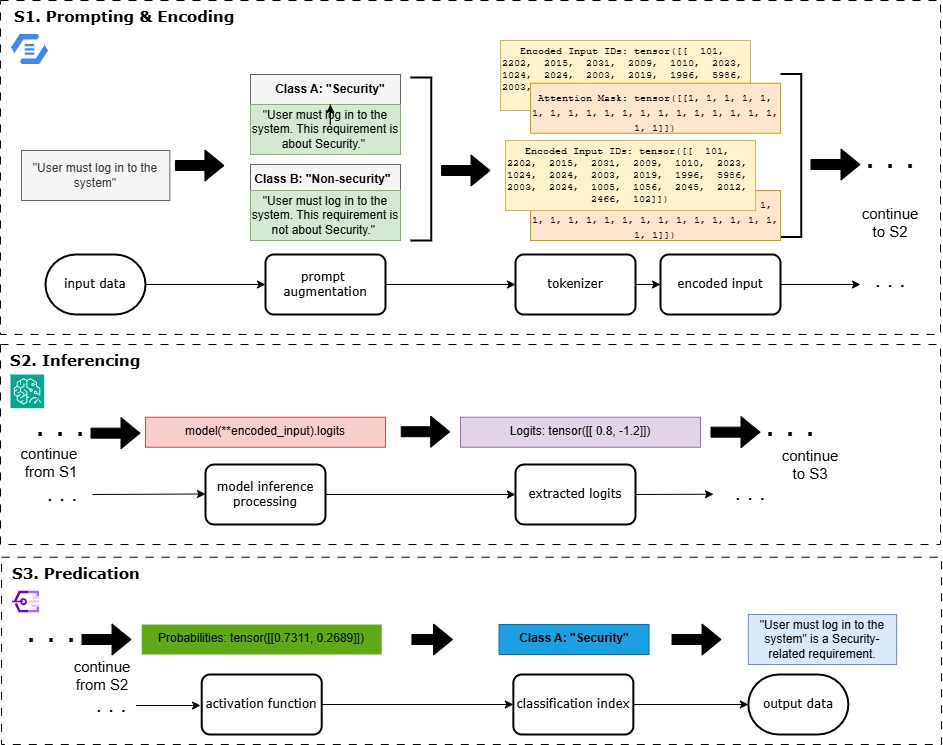}
    \caption{Inference-based learning for requirements classification with LLMs. }
    \label{fig:zsl-approach}
\end{figure}

\subsubsection{Model Prompting and Input Encoding}

The first step in applying generative LLMs to requirements classification is to craft appropriate prompts and encode the input data in a way that guides the model's response. This step ensures that the model is provided with the necessary context and instructions to effectively perform the task at hand.

This process involves several key activities:

\begin{enumerate}
    \item \textbf{Preparation of Input Data:} A batch of text data is prepared for classification, consisting of text samples paired with clear and structured instructions. These instructions explicitly specify the classification task and provide relevant context to help the model understand what is expected. The instructions often include a list of possible categories or labels that the model should consider when making predictions. For example, a prompt might be: \textit{“Classify the following text into one of these categories: [Label 1, Label 2, Label 3].”} This structure helps the model comprehend the task, particularly in zero-shot scenarios where the model has not been trained on specific classes but still needs to generate meaningful predictions.

    \item \textbf{Tokenisation:} Once the prompt is designed, the next step is tokenisation, where the input text and its associated instructions are divided into smaller units called tokens. These tokens are then mapped to unique identifiers based on a predefined vocabulary. Tokenisation is an essential step for transformer models, as it converts the input data into a format that the model can process.

    \item \textbf{Encoding into Numerical Representations:} After tokenisation, the tokens are transformed into numerical representations known as embeddings. These embeddings capture semantic information from the text, allowing the model to understand the relationships and context between tokens. By representing the input data in this format, the model can effectively analyse the content and make classifications.

    \item \textbf{Positional Encoding:} In transformer architectures, positional encodings are added to the embeddings to account for the sequential nature of the input data. These positional encodings ensure that the model can retain the order of the tokens, which is vital for understanding the meaning and relationships between tokens in a sequence. By incorporating positional information, the model can accurately process the input during its computations.

\end{enumerate}

\subsubsection{Inferencing}

In the inferencing step, the generative model processes the encoded inputs and applies its reasoning capabilities to make inferences based on the given requirements. The model then generates potential categories or labels that correspond to the input data.

After encoding the input data, including the prompts, it is passed through the transformer model. This model leverages multiple layers of attention mechanisms, including self-attention and feed-forward networks, which allow it to assess the importance of different tokens relative to one another. As the input data flows through these layers, the model generates an output tensor that contains intermediate results from the computations. These results represent the model's evolving understanding of the text and the associated classification task.

The transformer architecture's generative capabilities are central to its ability to capture complex patterns and relationships within the data. These capabilities make the model highly effective not only for classification tasks but also for more complex tasks such as text generation, summarisation, and reasoning. Following the inferencing process, the model outputs a set of \textbf{logits}—unnormalised scores that reflect the model's confidence in each potential classification. These logits are used to determine the most likely class or label for the input, based on the model's learned associations.

\subsubsection{Prediction}

In the final step, the model's inferences are transformed into a definitive classification prediction, assigning the input text to one or more relevant categories based on the inferencing results.

The extracted logits are processed through an activation function, typically \textbf{sigmoid} or \textbf{softmax}, to convert the raw prediction scores into probabilities. For binary ZSL, a \textbf{sigmoid} activation function (Equation \ref{eq:1}) is applied to the model's output, generating a probability score that reflects the likelihood of the target class being present. For multi-class ZSL, the \textbf{softmax} activation function (Equation \ref{eq:2}) is used, generating a probability distribution across all potential classes. In both cases, the log probabilities are exponentiated to produce probability values for each possible label. The class with the highest probability is then selected as the predicted category.

\begin{equation} \label{eq:1} 
sigmoid(z) = \frac{1}{1 + exp(-z)} 
\end{equation}

The sigmoid function takes a single input, denoted as $z$, and outputs a value between $0$ and $1$, which can be interpreted as the probability of the positive class in a binary classification problem.

\begin{equation} \label{eq:2} 
softmax(z_i) = \frac{exp(z_i)}{\sum_{j=1}^{K} exp(z_j)} 
\end{equation}

The softmax function processes the output of each neuron, denoted as $z_i$, exponentiates it, and computes the sum of the exponentials of all outputs in the denominator. The final output is the exponentiated value of the $i-th$ neuron, divided by the sum of all exponentials, yielding a probability distribution over the $K$ classes. The sum of all probabilities equals $1$, ensuring the model's predictions are proportionally distributed across the possible labels.

The classification index, or the selected label, is determined by identifying the label corresponding to the highest predicted probability. This index represents the model's predicted class for the input text, based on the learned representations and relationships established during training.

Together, these three stages—model prompting and input encoding, inferencing, and prediction—combine with the generative model's reasoning capabilities to form the basis for performing effective requirements classification using inference-based learning.

\subsection{Embedding-Based Learning}
\label{sec:EBL}

As discussed earlier, non-generative LLMs typically rely on embedding-based learning method for tasks such as requirements classification. In this study, we employ an embedding-based learning method for requirements classification, inspired by the approach outlined by Veeranna et al.~\cite{sappadla2016using}.

This method follows a structured process:

\begin{itemize}
    \item \textbf{Word Embeddings for Text and Class Labels:} Both the input text (i.e., the requirements) and the corresponding class labels (e.g., ``usability'', ``security'') are represented as word sequences using word embeddings. These embeddings capture the semantic relationships between words in a high-dimensional vector space.

    \item \textbf{Semantic Similarity Calculation:} Once the word embeddings for the input text and class labels are generated, the next step involves calculating the semantic similarity between the text sequence and each potential label sequence. This is done by computing similarity scores using predefined distance metrics, such as cosine similarity (see Equation \ref{eq:3}).

    \begin{equation}
\text{Cosine Similarity} = \cos(\theta) = \frac{\mathbf{A} \cdot \mathbf{B}}{\|\mathbf{A}\| \|\mathbf{B}\|}
\label{eq:3}
\end{equation}

\noindent where:
\begin{itemize}
    \item \(\mathbf{A}\) is the word embedding vector representing the input text sequence.
    \item \(\mathbf{B}\) is the word embedding vector representing a potential class label sequence.
    \item \(\|\mathbf{A}\|\) and \(\|\mathbf{B}\|\) are the magnitudes (norms) of the vectors \(\mathbf{A}\) and \(\mathbf{B}\), respectively.
    \item \(\mathbf{A} \cdot \mathbf{B}\) is the dot product of the vectors \(\mathbf{A}\) and \(\mathbf{B}\).
\end{itemize}

    \item \textbf{Classification:} If the similarity score between a given text and a label exceeds a predefined threshold, the text is classified into the category represented by that label. If no label exceeds the threshold, the text is not classified under any category.
\end{itemize}


It is important to note that the labels in this approach are treated as sequences of words, which can vary in length and may consist of either single words or multi-word phrases. This flexibility allows for a more complex and adaptable classification process.

To ensure that the labels are meaningful and contextually appropriate for requirements classification tasks, we have developed \textbf{expert-curated labels} based on our collective understanding of the various requirement classes. These labels were carefully crafted in our previous study~\cite{alhoshan2023zero}, and represent a comprehensive set of terms that are typically associated with specific classes of requirements. For example, the labels for the Functional Requirement (FR) class include terms such as ``functional'', ``system'', ``behavior'', ``shall'', and ``must'', which are commonly linked to functional requirements. In the Quality class, the label terms include ``quality'', ``performance'', ``efficiency'', and ``reliability''. In the current study, we will use these expert-curated labels as part of our embedding-based learning method to ensure a contextually accurate classification process.

\section{Research Methods}
\label{sec:methods}

\subsection{Research Questions}

The primary goal of our research is to assess the performance of generative LLMs in requirements classification across a diverse range of datasets, tasks, and prompt structures. In addition, we aim to compare their performance against that of non-generative LLMs when executing the same tasks. To achieve these objectives, we focus on addressing two key research questions:

\begin{itemize}

\item \textit{RQ1: How effective are generative LLMs in performing requirements classification? Specifically, how do different LLMs, dataset variations, task types, and prompt structures influence the classification performance?}

\item \textit{RQ2: How do generative LLMs perform in requirements classification compared to non-generative LLMs?}

\end{itemize}

RQ1 is based on the hypothesis that the diversity in datasets, tasks, and prompts significantly affects model performance in requirements classification. To test this, we will conduct a series of systematic experiments and assess the statistical significance of model performance across these factors.

RQ2 assumes that non-generative LLMs may outperform generative LLMs in the requirements classification domain, as classification tasks are generally considered a strength of non-generative models. To explore this hypothesis, we will compare the performance of generative and non-generative LLMs through controlled experiments. Additionally, we will assess the statistical significance of any observed differences to better understand the magnitude of their performance gap.

Below, we outline the experimental design for addressing both research questions (RQ1 and RQ2). This includes an overview of the essential components of our experiments—such as models, datasets, tasks, and prompts—along with the evaluation metrics, statistical methods used to assess model performance, and the experimental setup.

\subsection{Experimental Design for RQ1}
\label{sec:RQ1design}

\subsubsection{Models}
\label{sec:RQ1models}

We have selected three state-of-the-art generative LLMs for our experiments: BigScience's \textit{Bloom} \cite{le2023bloom,gemmamodelcard}, Google DeepMind's \textit{Gemma} \cite{team2024gemma, gemmamodelcard}, and Meta AI's \textit{Llama} \cite{touvron2023llama,llama3modelcard}. Below, we provide a brief introduction to these models, focusing on key aspects such as their training data, token and parameter sizes, and architecture. All these components are important to understand the model's ability to learn complex pattern, its computational efficiency, and its suitability for specific tasks. A summary of these details for each selected LLM is presented in Table \ref{tab:LLM} for easy reference. 

\begin{table}
\centering
\caption{Summary of Selected Generative LLMs for Our Experiments.}
\label{tab:LLM}
\resizebox{\textwidth}{!}{
\begin{tblr}{
  cells = {c},
  hline{1-2,5} = {-}{},
}
\textbf{LLMs} & \textbf{Developer} & {\textbf{Parameter~\textbf{and }}\\\textbf{\textbf{Token Size}}} & \textbf{Architecture}                                    & {\textbf{Attention~}\\\textbf{Mechanism~}} & \textbf{Training Data}                                               & {\textbf{Model Variants}\\\textbf{~from HuggingFace}} \\
Bloom         & BigScience         & {560m parameters\\(5.6B tokens)}                                 & {Transformer\\Decoder-only~}                             & Self-attention                                        & {diverse text: 45 languages and\\13 programming languages.}          & bigscience/bloom-560m                                 \\
Gemma         & Google             & {2B parameters\\(6T tokens)}                                     & {Transformer\\Decoder-only}                              & Multi-query                                         & {primarily-English data fromweb \\documents, mathematics, and code.} & google/gemma-2b                                       \\
Llama         & Meta               & {8B parameters~\\(15T tokens)}                                   & {Enhanced \\Transformer\\Decoder-only\\w. SFT and RLHF~} & Grouped-query                                        & {publicly available online data\\and 10m human labelled examples.}   & meta-llama/Meta-Llama-3-8B                            
\end{tblr}
}
\end{table}

\textit{Training Data:} All three LLMs--Bloom, Gemma, and Llama--leverage large-scale datasets to optimise performance, though their data sources \textit{ might differ}. Bloom was trained on 1.5TB of multilingual text (45 natural languages, predominantly English at 31.3\%) and 12 programming languages, emphasising linguistic diversity and code integration. Gemma prioritises breadth through web documents (primarily English) for diverse linguistic exposure, supplemented by code to grasp programming syntax and mathematical text to enhance logical reasoning. Llama uses publicly available data for pretraining, avoiding proprietary user data, and fine-tuned on public instruction datasets and over 10 million human-annotated examples to align outputs with human preferences for providing feedbacks. 
   
\textit{Token and Prameter size:} Bloom, with 559 million parameters, was trained on 5.6 billion tokens, sourced from multilingual text and programming languages codes. Gemma, a smaller model with 2 billion parameters, was trained on a much larger dataset of 6 trillion tokens, focusing on diverse web-based sources. Llama, with 8 billion parameters, was pretrained on an even larger dataset of 15 trillion tokens, sourced from publicly available data. These differences \textit {might show a trade-off} between model capacity (parameters) and data diversity/scale (tokens). Larger models with more parameters can capture more complex patterns, but they require massive datasets to fully utilize their capacity. Conversely, smaller models may be more efficient but might underperform on tasks requiring broad or deep understanding unless trained on sufficiently diverse and large datasets.
   
\textit{Architecture and Attention Mechanisms:} All three LLMs—Bloom, Gemma, and Llama—are based on the transformer architecture, which utilises the decoder-only design popularised by models like GPT \cite{dale2021gpt,achiam2023gpt}. This design, optimised for autoregressive text generation, enables these models to predict and generate coherent text by processing tokens sequentially. However, while all three LLMs share the foundational transformer decoder design, they differ in their attention mechanisms. Specifically, Bloom is based on the traditional \textit{self-attention} mechanism, Gemma adopts \textit{multi-query attention}, and Llama is based on \textit{grouped-query attention}. 

In our experiments, we will use an implementation version of these models provided by HuggingFace\footnote{https://huggingface.co/models}, as indicated in Table \ref{tab:LLM}.

\subsubsection{Datasets}
\label{sec:RQ1datasets}

We have selected the following three requirements datasets to evaluate model performance in our experiments, as these datasets have been widely used by researchers exploring machine learning in the context of RE. 

\begin{itemize}
    
    \item \textbf{PROMISE NFR Dataset:} This dataset was developed by Cleland-Huang \textit{et al.} \cite{cleland2007automated,jane_cleland_huang_2007_268542}, and has been widely used by researchers, e.g., by Kurtanovi{\'c} and Maalej \cite{kurtanovic2017automatically}, and by Hey \textit{et al.} \cite{hey2020norbert}. The dataset contains 625 requirements, consisting of 255 FRs and 370 NFRs. The NFRs are divided into 11 classes, what are: Usability (67 requirements), Security (66 requirements), Operational (62 requirements), Performance (54 requirements), Look \& feel (38 requirements), Availability (21 requirements), SC = Scalability (21 requirements), Maintainability (17 requirements), Legal (13 requirements), Fault tolerance (10 requirements), and PO = Portability (1 requirement). As the Portability class only contains one requirement, it has been excluded from our experiments. Consequently, our experiments will only consider 10 NFR classes plus one FR class (treating all FRs as one class).

    \item \textbf{Functional-Quality Dataset:} Developed by Dalpiaz \textit{et al.}~\cite{dalpiaz2019requirements}, this dataset is based on a re-annotation of the PROMISE NFR dataset, plus other open source and closed-source projects. The portion of the dataset used in our study includes the following project: the reclassification of PROMISE, Dronology, Wasp, Leeds, and ReqView, for a total of 956 requirements. The data considered for the binary classification task are classified into Quality (522) and non-Quality (434); and Functional (587) and non-Functional (387).

    \item \textbf{SecReq Dataset:} Created by Knauss \textit{et al.} \cite{knauss2011supporting,knausseric20214530183}, this dataset contains 510 requirements, made of security-related requirements (187) and non-security related requirements (323). The requirements were collected from three projects: Common Electronic Purse (ePurse), Customer Premises Network (CPN), and Global Platform Spec (GPS). The dataset has been used, e.g., by Varenov \textit{et al.}~\cite{varenov2021security}.

\end{itemize}

\subsubsection{Tasks}
\label{sec:RQ1tasks}

Based on the selected datasets, we have defined five requirements classification tasks for evaluation in our experiments, including three binary classification tasks and two multi-class classification tasks. These tasks, along with their corresponding datasets, are described below:

\textit{Tasks on PROMISE NFR Dataset:}

    \begin{itemize}
        \item \textbf{Task NFR}: This task performs multi-class classification by assigning each requirement into one of the aforementioned 10 NFR classes in PROMISE NFR Dataset. 
        \item \textbf{Task NFR-Top4}: This task also performs multi-class classification but only considers the four largest classes in PROMISE NFR Dataset, comprising Usability (67 requirements), Security (66 requirements), Operational (62 requirements), and Performance (54 requirements).   
    \end{itemize}

\textit{Tasks on Functional-Quality Dataset:}

    \begin{itemize}
        \item  \textbf{Task Functional}: This task performs binary classification to distinguish between functional and non-functional requirements, assuming that a requirement in the dataset belongs to either a functional or a non-functional requirements class. 
        \item \textbf{Task Quality}: This binary classification task distinguishes between the quality requirements and non-quality requirements.  
    \end{itemize}

\textit{Task on SeqReq Dataset:}

    \begin{itemize}
        \item \textbf{Task Security}: This binary classification task distinguishes between the security related and the non-security related requirements. 
    \end{itemize}

\subsubsection{Prompts}
\label{sec:prompt}

To conduct a thorough investigation of the selected LLMs, we have carefully designed six \textbf{prompt patterns} to assess which prompts are most effective in guiding the LLMs through classification tasks. These prompt patterns are organized into three groups: \textit{``Assertion-Based,'' ``Definition-Based,''} and \textit{``Q/A-Based,''} which are described below.

\textbf{Assertion-Based Prompts:} These prompts present an assertive statement to the LLMs. We consider two assertion-based prompt patterns: ``is about'' and ``belongs to.'' The \textit{``is about''} pattern treats requirements classification as a topic modeling task, guiding the LLM to determine whether a requirement pertains to a specific topic. On the other hand, the \textit{``belongs to''} pattern directs the LLM to assess whether a given requirement statement belongs to a specific type of requirement. Example prompts include: ``This requirement is about usability'' and ``This requirement belongs to usability.'' We refer to these two prompt patterns as the \textit{``is-about assertion''} and \textit{``belongs-to assertion''}. 

\textbf{Definition-Based Prompts:} These prompts build upon the assertion-based prompts by incorporating the definition of a specific type of requirement as a prefix. The LLM is then asked to infer whether a given requirement \textit{``is about''} or \textit{``belongs to''} the defined type of requirement. For instance, an example of the \textit{``is about''} definition-based prompt is: ``Usability requirements are quality requirements that define what a system must do to support users' task performance. Therefore, this requirement is about usability.'' Similarly, an example of the \textit{``belongs to''} definition-based prompt is: ``Usability requirements are quality requirements that define what a system must do to support users' task performance. Therefore, this requirement belongs to usability.'' We refer to these two prompt patterns as the \textbf{``is-about definition''} and \textit{``belongs-to definition''}.

\textbf{Q/A-Based Prompts:} These prompts are the question-based variation of the \textbf{``Assertion-Based Prompts.''} In this case, the LLM is asked to respond with a yes or no answer to whether a given requirement \textit{``is about''} or \textit{``belongs to''} a specific type of requirement, based on the provided class label. Example prompts include: ``Is this requirement about usability?'' and ``Does this requirement belong to usability?'' We refer to these two prompt patterns as the \textit{``is-about Q/A''} and \textit{``belongs-to Q/A''}.

\subsubsection{Dataset Variations}
\label{sec:RQ1variation}

To assess whether variations within a dataset can impact the performance of LLMs in requirements classification, we apply the following five modifications to each dataset:

\begin{itemize}
    \item \textit{Modifications to requirement texts:} 
    1) Punctuation removal (remove all punctuation marks, such as commas, full stops, and colons, from the requirement texts); 
    2) Sentence completion (ensure each requirement statement ends with a full stop).
    
    \item \textit{Modifications to class labels:} 
    1) Class labels with only lowercase letters; 
    2) Class labels with only uppercase letters; 
    3) Class labels with the first letter capitalised. 
\end{itemize}

Our experiments will consider these modifications as five dataset variations. 

\subsubsection{Evaluation Metrics}
\label{sec:RQ1metrics}

The following standard metrics have been selected to evaluate the performance of each LLM in our experiments:

\begin{itemize}
    \item \textit{Performance on individual classes:} We will use standard, unweighted \textit{precision ($P$), recall ($R$), and their weighted harmonic mean, $F1$}, to assess the model's performance on individual classes in the dataset. These metrics will provide three scores for each class in the dataset.
    
    \item \textit{Performance across all classes:} We will use weighted \textit{precision ($wP$), recall ($wR$), and F1-score ($wF1$)} to evaluate the overall model performance across all classes in the dataset. These metrics will yield three weighted average scores, which represent the mean of all per-class $P$ scores, all per-class $R$ scores, and all per-class $F1$ scores, while accounting for each class's support (i.e., the number of requirements in that class).
\end{itemize}

\subsubsection{Statistical Analysis}
\label{sec:RQ1test}

To assess the statistical significance of the factors affecting the performance of requirements classification, we consider these factors in our experiments: \textbf{LLMs, dataset variations, prompt patterns,} and \textbf{task types}. We apply a \textbf{repeated measures design}, which allows us to test the same datasets under different combinations of these factors. This design enables direct comparison of the effects of each factor on the dataset, minimising variability due to differences between datasets. By testing all combinations of factors on the same dataset, we can more accurately evaluate how each factor impacts performance.

To analyse the effect of each factor individually, we organise the results to isolate each factor. For instance, to analyse the impact of the LLM factor, we compare the results of Bloom, Gemma, and Llama, even though these results are derived from combinations of all factors. We primarily evaluate performance using the weighted F1 score ($wF1$), and repeat the tests to separately examine weighted precision ($wP$) and weighted recall ($wR$) for each factor. This approach enables a comprehensive understanding of how each factor affects $F1$, $wP$, and $wR$ scores.

To determine the statistical significance of the performance results (i.e., the $wP$, $wR$, and $wF1$ scores), we will perform the \textit{Friedman Test} \cite{friedman1937use} on each factor. The Friedman test is a non-parametric statistical method that compares the rank order of model performance under each condition (or factor), rather than comparing the performance scores directly. Instead of relying on exact numerical scores, the test ranks the models (e.g., $1^{st}$, $2^{nd}$, $3^{rd}$) based on their performance relative to others, making it easier to compare factors and identify which ones lead to better or worse performance, even when the raw scores are close or vary slightly. This approach is particularly beneficial for complex or non-uniform data, such as the performance results of $wP$, $wR$, and $wF1$, ensuring more reliable and interpretable results.

For each factor under investigation ($\sigma \in \{LLMs, dataset ~variations, prompt patterns, task types\}$), we state two hypotheses:

\begin{description}

\item[Null Hypothesis ($H_0$):] The factor $\sigma$ \textit{does not significantly affect} model performance in requirements classification.

\item[Alternative Hypothesis ($H_1$):] The factor $\sigma$ \textit{significantly affects} model performance in requirements classification. 

\end{description}

The Friedman test will generate a $p$-value for each factor, which we will use to test the hypotheses. Specifically, if the $p$-value is $\leq 0.05$, we reject the null hypothesis ($H_0$) and accept the alternative hypothesis ($H_1$), indicating that the factor ($\sigma$) has a significant impact on performance. Conversely, if the $p$-value is $> 0.05$, we fail to reject the null hypothesis, meaning there is no significant effect of the factor on model performance\footnote{$p = 0.05$ is commonly used as a threshold for statistical significance in hypothesis testing}.

\subsubsection{Experimental Setup for RQ1}
\label{sec:RQ1exp}

Based on the aforementioned four factors, we have designed two groups of experiments for each selected LLM:

\begin{itemize}
\item For binary classification, there will be a total of 90 experimental settings (\textit{= 6 prompts $\times$ 5 dataset variations $\times$ 3 tasks}).
\item For multi-class classification, there will be a total of 60 experimental settings (\textit{= 6 prompts $\times$ 5 dataset variations $\times$ 2 tasks}).
\end{itemize}

Thus, to investigate each LLM, we need to conduct 90 + 60 = 150 experiments. For all three LLMs, the total number of experiments is 3 $times$ 150 = 450. We set up these experiments on Google Colab, to enable better collaboration and code sharing.

Each model is evaluated using an inference-based learning approach, where the LLM is tasked with generating predictions based on its learned knowledge, without the need for explicit retraining. Inference-based learning leverages the pre-trained capabilities of the LLM to adapt to new tasks through prompt engineering and task-specific input, allowing the model to generalise to various datasets and tasks. This approach described in detail in Section \ref{sec:approach}.

\subsection{Experimental Design for RQ2}
\label{sec:RQ2design}

\subsubsection{Baseline Models}
\label{sec:RQ2models}

To address RQ2, we will use two non-generative LLMs—\textbf{SBERT} and \textbf{All-Mini}—as baseline models to compare the performance of the best-performing LLMs identified in our RQ1 experiments. These non-generative models were selected because they produced the best performance results in our preliminary experiments \cite{alhoshan2023zero}. Specifically, SBERT achieved the highest performance for the binary classification task \textbf{Task Functional}, while All-Mini outperformed SBERT in the binary classification task \textbf{Task Security} and in the multi-class classification task \textbf{Task NFR}.

\subsubsection{Experimental Setup for RQ2} 

In our previous study \cite{alhoshan2023zero}, SBERT and All-Mini were used to perform three classification tasks: two binary classification tasks—\textbf{Task Functional} and \textbf{Task Security}—and one multi-class classification task, \textbf{Task NFR}. For the current study, which involves five tasks, we will apply SBERT and All-Mini to perform these same tasks. As such, we have designed five experiments for each model, with one experiment for each task.

Each model is evaluated for each task using an embedding-based learning approach, in which the model generates predictions based on its pre-trained knowledge without requiring explicit retraining. Embedding-based learning utilises the model's ability to map input data into high-dimensional representations (embeddings), which are then used to make task-specific predictions. By leveraging these embeddings, both SBERT and All-Mini can effectively adapt to new tasks through task-specific input and expert-curated labels, making them highly flexible for generalising across various datasets and classification tasks. 

The same datasets and evaluation metrics used in the RQ1 experiments will be applied to evaluate the performance of SBERT and All-Mini. However, in this phase of evaluation, different dataset variations will not be considered.

\subsubsection{Statistical Analysis}

To determine whether the performance differences between the best generative LLMs and the best non-generative LLMs are statistically significant, we will apply the \textit{Wilcoxon Signed-Rank Test} \cite{wilcoxon1945individual}. This non-parametric test is ideal for pairwise comparisons and can handle non-uniformly distributed performance scores. We focus on $wF1$ as the primary evaluation metric, as it provides a balanced assessment of both $wP$ and $wR$, making it especially useful for imbalanced datasets.

In our analysis, we compare the best-performing model from each approach as follows: First, we divide the set of requirement statements from each dataset and task into groups. We based our experiments on grouping 3–4 requirement statements together to properly perform the test. Next, we calculate group-level $wF1$ scores for each group, taking into account the class distributions within each group to balance $P$ and $R$.

This process generates two sets of $wF1$ scores, one for each approach. These scores are then ranked for each approach. Finally, the Wilcoxon Signed-Rank Test is applied to each classification task, allowing us to assess whether there is a statistically significant difference in performance between the two approaches, both for binary classification tasks (Functional, Quality, and Security) and multi-class classification tasks (NFR and NFR-Top4).

The hypotheses for each of the five classification tasks are as follows:

\begin{description}
    \item[Null Hypothesis ($H_0$):] There is \textit{no significant difference} in performance between the best generative LLM and the best non-generative LLM for the designated task.
    
    \item[Alternative Hypothesis ($H_1$):] There is a \textit{significant difference} in performance between the best generative LLM and the best non-generative LLM for the designated task.
\end{description}

The Wilcoxon Signed-Rank Test will generate a $p$-value for each task. In hypothesis testing, the $p$-value helps determine the significance of the performance results. Specifically, it indicates the probability of observing the test results under the assumption that the null hypothesis is true:

\begin{itemize}
    \item If \textbf{the $p$-value is small (typically $\leq 0.05$)}, we reject the null hypothesis ($H_0$) in favor of the alternative hypothesis ($H_1$), indicating that there is sufficient evidence to conclude that there is a \textbf{significant difference} between the performance of the two models.
    
    \item If \textbf{the $p$-value is large (typically $> 0.05$)}, we fail to reject the null hypothesis ($H_0$), implying that there is no significant performance difference between the two models, and the observed difference may be due to random variation.
\end{itemize}

\section{Experimental Results for RQ1}
\label{sec:res4RQ1}

In this section, we present and analyze the experimental results addressing the first research question (RQ1): \textit{How effective are generative LLMs in performing requirements classification? Specifically, how do different LLMs, dataset variations, task types, and prompt structures influence classification performance?} We begin by evaluating the impact of these factors on classification performance, followed by a detailed analysis of their specific effects. Next, we apply the Friedman test to identify which factors significantly influence performance. Finally, we synthesize the findings to provide a comprehensive answer to the research question.


\subsection{Impact of Generative LLMs}

\begin{table}
\centering
\caption{Impact of Dataset Variations, Tasks and Prompt Patterns on LLMs. It also highlights how frequently each LLM is affected by these factors.}
\label{tab:LLM-factor-analysis}
\scriptsize
\begin{tblr}{
  row{1} = {c},
  row{2} = {c},
  row{3} = {c},
  row{14} = {c},
  row{15} = {c},
  row{16} = {c},
  cell{1}{1} = {c=5}{},
  cell{2}{1} = {r=2}{},
  cell{2}{2} = {r=2}{},
  cell{2}{3} = {c=3}{},
  cell{4}{1} = {r=5}{c},
  cell{4}{3} = {c},
  cell{4}{4} = {c},
  cell{4}{5} = {c},
  cell{5}{3} = {c},
  cell{5}{4} = {c},
  cell{5}{5} = {c},
  cell{6}{3} = {c},
  cell{6}{4} = {c},
  cell{6}{5} = {c},
  cell{7}{3} = {c},
  cell{7}{4} = {c},
  cell{7}{5} = {c},
  cell{8}{3} = {c},
  cell{8}{4} = {c},
  cell{8}{5} = {c},
  cell{9}{1} = {r=5}{c},
  cell{9}{3} = {c},
  cell{9}{4} = {c},
  cell{9}{5} = {c},
  cell{10}{3} = {c},
  cell{10}{4} = {c},
  cell{10}{5} = {c},
  cell{11}{3} = {c},
  cell{11}{4} = {c},
  cell{11}{5} = {c},
  cell{12}{3} = {c},
  cell{12}{4} = {c},
  cell{12}{5} = {c},
  cell{13}{3} = {c},
  cell{13}{4} = {c},
  cell{13}{5} = {c},
  cell{14}{1} = {c=5}{},
  cell{15}{1} = {r=2}{},
  cell{15}{2} = {r=2}{},
  cell{15}{3} = {c=3}{},
  cell{17}{1} = {r=5}{c},
  cell{17}{3} = {c},
  cell{17}{4} = {c},
  cell{17}{5} = {c},
  cell{18}{3} = {c},
  cell{18}{4} = {c},
  cell{18}{5} = {c},
  cell{19}{3} = {c},
  cell{19}{4} = {c},
  cell{19}{5} = {c},
  cell{20}{3} = {c},
  cell{20}{4} = {c},
  cell{20}{5} = {c},
  cell{21}{3} = {c},
  cell{21}{4} = {c},
  cell{21}{5} = {c},
  cell{22}{1} = {r=5}{c},
  cell{22}{3} = {c},
  cell{22}{4} = {c},
  cell{22}{5} = {c},
  cell{23}{3} = {c},
  cell{23}{4} = {c},
  cell{23}{5} = {c},
  cell{24}{3} = {c},
  cell{24}{4} = {c},
  cell{24}{5} = {c},
  cell{25}{3} = {c},
  cell{25}{4} = {c},
  cell{25}{5} = {c},
  cell{26}{3} = {c},
  cell{26}{4} = {c},
  cell{26}{5} = {c},
  hline{1-2,4,14-15,17,22,27} = {-}{},
  hline{3,16} = {3-5}{},
  hline{7,12,20,25} = {2-5}{dashed},
  hline{9} = {2-5}{},
}
\textbf{\textbf{A) Performance of Generative LLMs on Binary Classification Tasks}}    &                                                       &                &                &                \\
\textbf{\textbf{Metrics}}                                                             & \textbf{\textbf{Factors affecting Model Performance}} & \textbf{LLMs } &                &                \\
                                                                                      &                                                       & \textbf{Bloom} & \textbf{Gemma} & \textbf{Llama} \\
\textbf{$wP$}                                                                         & Dataset Variations (5 Experiments)                    & 3/5            & 0/5            & 2/5            \\
                                                                                      & Prompt Patterns (6 Experiments)                       & 4/6            & 0/6            & 2/6            \\
                                                                                      & Tasks (3 Experiments)                                 & 2/3            & 1/3            & 0/3            \\
                                                                                      & \textit{Max. $wP$ score}                              & 0.7683         & 0.7611         & 0.7254         \\
                                                                                      & \textit{Avg. $wP$ score}                              & 0.42           & 0.50           & 0.50           \\
~\textbf{$wR$}                                                                        & Dataset Variation (5 Experiments)                     & 0/5            & 4/5            & 1/5            \\
                                                                                      & Prompt Patterns  (6 Experiments)                      & 0/6            & 3/6            & 3/6            \\
                                                                                      & Tasks (3 Experiments)                                 & 0/3            & 2/3            & 1/3            \\
                                                                                      & \textit{Max. $wR$ score~}                             & 0.6046         & 0.6588         & 0.6527         \\
                                                                                      & \textit{Avg. $wR$ score~}                             & 0.50           & 0.51           & 0.49           \\
\textbf{\textbf{B) Performance of~}\textbf{LLMs on Multi-Class Classification Tasks}} &                                                       &                &                &                \\
\textbf{\textbf{Metrics}}                                                             & \textbf{\textbf{Factors affecting Model Performance}} & \textbf{LLMs } &                &                \\
                                                                                      &                                                       & \textbf{Bloom} & \textbf{Gemma} & \textbf{Llama} \\
\textbf{$wP$}                                                                         & Dataset Variations (5 Experiments)                    & 4/5            & 0/5            & 1/5            \\
                                                                                      & Prompt Patterns (6 Experiments)                       & 3/6            & 1/6            & 2/6            \\
                                                                                      & Tasks (2 Experiments)                                 & 1/2            & 0/2            & 1/2            \\
                                                                                      & \textit{Max. $wP$ score}                              & 0.5999         & 0.4048         & 0.4103         \\
                                                                                      & \textit{Avg. $wP$ score}                              & 0.14           & 0.07           & 0.15           \\
\textbf{$wR$}                                                                         & Dataset variations (5 Experiments)                    & 0/5            & 1/5            & 4/5            \\
                                                                                      & Prompt patterns~(6 Experiments)                       & 3/6            & 0/6            & 3/6            \\
                                                                                      & Tasks~(2 Experiments)                                 & 0/2            & 0/2            & 2/2            \\
                                                                                      & \textit{Max. $wR$ score}                              & 0.3092         & 0.2489         & 0.4257         \\
                                                                                      & \textit{Avg. $wR$ score}                              & 0.18           & 0.11           & 0.17           
\end{tblr}
\end{table}

Table \ref{tab:LLM-factor-analysis} presents the performance of the three LLMs across both binary and multi-class classification tasks. It also highlights the number of times each LLM outperforms the others across different dataset variations, prompts, and tasks. Below, we provide a detailed analysis of these results.

\subsubsection{Model Performance on Binary Classification}

The performance of the three LLMs on three binary classification tasks reveals the following:

\begin{itemize}
    \item \textbf{Weighted Precision:} When measured by weighted precision, Bloom achieved the highest precision score ($Max.wP=0.7683$), but fell short in terms of average precision. In contrast, Gemma and Llama jointly achieved the best average precision score ($Avg.wP=0.50$). Across the experimental factors, Bloom outperformed the other models in 3 out of 5 experiments related to dataset variations, 4 out of 6 experiments regarding prompt patterns, and 2 out of 3 experiments concerning tasks. Gemma performed the worst in terms of dataset variations and prompt patterns, while Llama was the least effective across the tasks.

    \item \textbf{Weighted Recall:} However, when measured by weighted recall, the results shifted. Gemma surpassed the other models, achieving both the highest recall score ($Max.wR=0.6588$) and the highest average recall ($Avg.wR=0.51$). Against the three experimental factors, Gemma led in 4 out of 5 experiments on dataset variations, 3 out of 6 experiments on prompt patterns, and 2 out of 3 tasks. In comparison, Bloom performed poorly across all three experimental factors.
\end{itemize}

Overall, Llama exhibited the most consistency and stability, delivering a balanced performance across both weighted precision and weighted recall. Gemma also demonstrated strong, reliable performance across both metrics. Based on the results presented in Table \ref{tab:LLM-factor-analysis}, we conclude that Llama and Gemma are the top-performing models for binary classification tasks, sharing the distinction of joint best models.

\subsubsection{Model Performance on Multi-Class Classification}

The performance of the three LLMs on the two multi-class classification tasks reveals the following:

\begin{itemize}
    \item \textbf{Weighted Precision:} When measured by weighted precision, Bloom achieved the highest precision score ($Max.wP=0.5999$), while Llama achieved the best average precision score ($Avg.wP=0.15$). Across the experimental factors, Bloom outperformed the other models in 4 out of 5 experiments related to dataset variations, 3 out of 6 experiments regarding prompt patterns, and 1 out of 2 experiments concerning tasks. Gemma performed the worst in terms of dataset variations and tasks.

    \item \textbf{Weighted Recall:} However, when measured by weighted recall, Llama achieved the highest recall score ($Max.wR=0.4257$), while Bloom achieved the best average recall score ($Avg.wR=0.18$). Against the three experimental factors, Llama led in 4 out of 5 experiments on dataset variations, 3 out of 6 experiments on prompt patterns, and 2 out of 2 tasks. In comparison, Gemma performed poorly across all three experimental factors.
\end{itemize}

Overall, Llama exhibited the most consistency and stability, delivering a balanced performance across both weighted precision and weighted recall. Bloom also demonstrated consistent performance across both metrics. Based on the results presented in Table \ref{tab:LLM-factor-analysis}, we conclude that Llama is the top-performing model for multi-class classification tasks.

\subsubsection{Key Findings and Insights}

The key results derived from our model performance experiments are summarised below. 

\begin{tcolorbox}[colback=green!5!white,colframe=green!75!black,title=Key Findings on LLMs]\label{RQ1LLMBox1}
\small
  \begin{itemize} 
    \item \textbf{Bloom} showed strong performance in weighted precision for both binary and multi-class classification tasks, but its performance in weighted recall was weaker, particularly for binary classification, with only slight improvement in multi-class recall. 
    \item \textbf{Gemma} excelled in binary classification, especially in weighted recall, but struggled significantly with multi-class classification tasks. 
    \item \textbf{Llama} demonstrated outstanding performance in multi-class classification, particularly in weighted recall, while maintaining balanced results in binary classification tasks. 
  \end{itemize}
\end{tcolorbox}

An important finding from our experiments is that \textit{the choice of LLM significantly impacts the outcomes of requirement classification tasks}. This is evident in the number of times each LLM achieved the best performance, as well as the differences observed in average and maximum scores when evaluating the weighted metrics \(wP\) and \(wR\). Further analysis reveals that some LLMs excel in binary classification tasks, while others perform better in multi-class classification scenarios. This suggests that the optimal LLM choice is likely \textit{task-dependent}, with different LLMs exhibiting strengths aligned with specific classification complexities.

For example, \textbf{Bloom} demonstrates notable performance in \(wP\) for both binary and multi-class classification tasks compared to the other LLMs, but it struggles with \(wR\), particularly in binary classification. This observation suggests that Bloom has a bias toward minimising false positives at the cost of increasing false negatives. Its transformer-decoder-only architecture, with self-attention mechanisms, excels at capturing complex linguistic patterns, contributing to accurate positive predictions (high precision). Although Bloom's multilingual training dataset is substantial, it is smaller than those used to train Gemma and Llama. However, its relatively large, multilingual training data, which promotes generalisation, may introduce ambiguity. This could lead to a conservative classification threshold, especially if certain classes are underrepresented in common languages or have more varied expressions across languages. Bloom's cautious approach, combined with its ability to detect complex patterns in diverse training data, results in fewer positive predictions, missing many true positives and thereby reducing recall. Essentially, Bloom prioritises certainty in positive predictions, potentially sacrificing the identification of less clear-cut, but still valid, positive instances.

\textbf{Gemma}, on the other hand, excels in binary classification, particularly in identifying rare positive instances, which results in a high weighted recall (\(wR\)). This makes it suitable for applications where missing positive cases is costly. Its strong performance in binary classification likely stems from its massive and diverse training data. However, Gemma struggles with multi-class classification. While its attention mechanism (i.e., multi-task attention) may improve efficiency, it might not sufficiently address the complex semantic distinctions required for accurate multi-class classification. Essentially, multi-query attention aids with speed and location but does not necessarily improve understanding of the subtle differences between multiple classes. Other factors, such as training data balance, may also contribute to this limitation.

\textbf{Llama} excels in multi-class classification, particularly in weighted recall (\(wR\)), likely due to its optimised transformer architecture, which enhances its ability to discern complex differences between multiple classes. Its extensive pre-training on 15 trillion tokens of public data, combined with techniques like Supervised Fine-Tuning (SFT) and Reinforcement Learning from Human Feedback (RLHF), further refines its understanding of diverse textual patterns and aligns its outputs with human preferences, improving its ability to identify all instances of various classes (high recall). This alignment, along with its efficient attention mechanism (i.e., grouped-query attention) for handling longer contexts, likely contributes to its balanced performance in binary classification as well. These factors enable Llama to achieve high recall in multi-class tasks while maintaining a balanced performance in binary classification, effectively balancing precision and recall across diverse tasks.

We summarise the key insights and observations gained from our research as follows:

\begin{tcolorbox}[colback=green!5!white,colframe=green!75!black,title=Key Insights into LLMs]\label{RQ1LLMBox2}
\small
  \begin{itemize}
         \item \textbf{Bloom's} architecture, which uses self-attention, promotes high precision in both binary and multi-class classification tasks. However, its relatively smaller training dataset compared to Gemma and Llama, along with potential ambiguities in its multilingual data, results in a more conservative classification threshold, negatively impacting recall, particularly in binary tasks.
        \item \textbf{Gemma's} architecture, incorporating an efficient multi-query attention mechanism, excels at binary classification, especially in recall. However, it struggles to capture the necessary semantic distinctions for accurate multi-class classification, limiting its performance in such tasks.
        \item \textbf{Llama's} optimised transformer architecture, combined with SFT and RLHF techniques and its efficient grouped-query attention mechanism, along with extensive pre-training and human feedback alignment, enables strong performance in multi-class classification, particularly in recall, by effectively discerning complex differences between classes. This architecture also contributes to balanced performance in binary classification tasks.
  \end{itemize}
\end{tcolorbox}

\subsection{Impact of Dataset Variations}

\begin{table}
\centering
\caption{Impact of Data Variations---Punctuation Removal (punct), Sentence Completion (sent), Lowercase (Lcase), Uppercase (Ucase), and Capitalization (Cap)---on LLMs, Tasks and Prompt Patterns. It also highlights how frequently each data variation is positively affected by these factors.}
\label{tab:text-factor-analysis}
\scriptsize 
\begin{tblr}{
  row{1} = {c},
  row{2} = {c},
  row{3} = {c},
  row{14} = {c},
  row{15} = {c},
  row{16} = {c},
  cell{1}{1} = {c=7}{},
  cell{2}{1} = {r=2}{},
  cell{2}{2} = {r=2}{},
  cell{2}{3} = {c=5}{},
  cell{4}{1} = {r=5}{c},
  cell{4}{3} = {c},
  cell{4}{4} = {c},
  cell{4}{5} = {c},
  cell{4}{6} = {c},
  cell{4}{7} = {c},
  cell{5}{3} = {c},
  cell{5}{4} = {c},
  cell{5}{5} = {c},
  cell{5}{6} = {c},
  cell{5}{7} = {c},
  cell{6}{3} = {c},
  cell{6}{4} = {c},
  cell{6}{5} = {c},
  cell{6}{6} = {c},
  cell{6}{7} = {c},
  cell{7}{3} = {c},
  cell{7}{4} = {c},
  cell{7}{5} = {c},
  cell{7}{6} = {c},
  cell{7}{7} = {c},
  cell{8}{3} = {c},
  cell{8}{4} = {c},
  cell{8}{5} = {c},
  cell{8}{6} = {c},
  cell{8}{7} = {c},
  cell{9}{1} = {r=5}{c},
  cell{9}{3} = {c},
  cell{9}{4} = {c},
  cell{9}{5} = {c},
  cell{9}{6} = {c},
  cell{9}{7} = {c},
  cell{10}{3} = {c},
  cell{10}{4} = {c},
  cell{10}{5} = {c},
  cell{10}{6} = {c},
  cell{10}{7} = {c},
  cell{11}{3} = {c},
  cell{11}{4} = {c},
  cell{11}{5} = {c},
  cell{11}{6} = {c},
  cell{11}{7} = {c},
  cell{12}{3} = {c},
  cell{12}{4} = {c},
  cell{12}{5} = {c},
  cell{12}{6} = {c},
  cell{12}{7} = {c},
  cell{13}{3} = {c},
  cell{13}{4} = {c},
  cell{13}{5} = {c},
  cell{13}{6} = {c},
  cell{13}{7} = {c},
  cell{14}{1} = {c=7}{},
  cell{15}{1} = {r=2}{},
  cell{15}{2} = {r=2}{},
  cell{15}{3} = {c=5}{},
  cell{17}{1} = {r=5}{c},
  cell{17}{3} = {c},
  cell{17}{4} = {c},
  cell{17}{5} = {c},
  cell{17}{6} = {c},
  cell{17}{7} = {c},
  cell{18}{3} = {c},
  cell{18}{4} = {c},
  cell{18}{5} = {c},
  cell{18}{6} = {c},
  cell{18}{7} = {c},
  cell{19}{3} = {c},
  cell{19}{4} = {c},
  cell{19}{5} = {c},
  cell{19}{6} = {c},
  cell{19}{7} = {c},
  cell{20}{3} = {c},
  cell{20}{4} = {c},
  cell{20}{5} = {c},
  cell{20}{6} = {c},
  cell{20}{7} = {c},
  cell{21}{3} = {c},
  cell{21}{4} = {c},
  cell{21}{5} = {c},
  cell{21}{6} = {c},
  cell{21}{7} = {c},
  cell{22}{1} = {r=5}{c},
  cell{22}{3} = {c},
  cell{22}{4} = {c},
  cell{22}{5} = {c},
  cell{22}{6} = {c},
  cell{22}{7} = {c},
  cell{23}{3} = {c},
  cell{23}{4} = {c},
  cell{23}{5} = {c},
  cell{23}{6} = {c},
  cell{23}{7} = {c},
  cell{24}{3} = {c},
  cell{24}{4} = {c},
  cell{24}{5} = {c},
  cell{24}{6} = {c},
  cell{24}{7} = {c},
  cell{25}{3} = {c},
  cell{25}{4} = {c},
  cell{25}{5} = {c},
  cell{25}{6} = {c},
  cell{25}{7} = {c},
  cell{26}{3} = {c},
  cell{26}{4} = {c},
  cell{26}{5} = {c},
  cell{26}{6} = {c},
  cell{26}{7} = {c},
  hline{1-2,4,9,14-15,17,22,27} = {-}{},
  hline{3,16} = {3-7}{},
  hline{7,12,20,25} = {2-7}{dashed},
}
\textbf{A) Effects of Dataset Variations on Other Factors in Binary Classification Setting}      &                                                 &                             &                &                &                &              \\
\textbf{Metrics }                                                                                & \textbf{Factors affected by Dataset Variations} & \textbf{Dataset Variations} &                &                &                &              \\
                                                                                                 &                                                 & \textbf{Punct.}             & \textbf{Sent.} & \textbf{Lcase} & \textbf{Ucase} & \textbf{Cap} \\
$wP$                                                                                             & LLMs~(3 Experiments)                            & 0/3                         & 1/3            & 0/3            & 0/3            & 2/3          \\
                                                                                                 & Prompt Patterns~(6 Experiments)                 & 0/6                         & 1/6            & 0/6            & 1/6            & 4/6          \\
                                                                                                 & Tasks~(3 Experiments)                           & 0/3                         & 0/3            & 0/3            & 0/3            & 3/3          \\
                                                                                                 & \textit{Max. $wP$ score}                        & 0.7254                      & 0.6902         & 0.7680         & 0.7680         & 0.7683       \\
                                                                                                 & \textit{Avg. ~$wP$score}                        & 0.47                        & 0.48           & 0.47           & 0.47           & 0.48         \\
$wR$                                                                                             & LLMs~(3 Experiments)                            & 1/3                         & 0/3            & 2/3            & 0/3            & 0/3          \\
                                                                                                 & Prompt Patterns~(6 Experiments)                 & 1/6                         & 1/6            & 3/6            & 1/6            & 0/6          \\
                                                                                                 & Tasks~(3 Experiments)                           & 0/3                         & 0/3            & 3/3            & 0/3            & 0/3          \\
                                                                                                 & \textit{Max. $wR$~score}                        & 0.6454                      & 0.6314         & 0.6588         & 0.6333         & 0.6490       \\
                                                                                                 & \textit{Avg. ~$wR$score}                        & 0.50                        & 0.50           & 0.50           & 0.50           & 0.50         \\
\textbf{B) Effects of Dataset Variations on Other Factors in Multi-class Classification Setting} &                                                 &                             &                &                &                &              \\
\textbf{Metrics}                                                                                 & \textbf{Factors affected by Dataset Variations} & \textbf{Dataset Variations} &                &                &                &              \\
                                                                                                 &                                                 & \textbf{Punct.}             & \textbf{Sent.} & \textbf{Lcase} & \textbf{Ucase} & \textbf{Cap} \\
$wP$                                                                                             & LLMs~(3 Experiments)                            & 0/3                         & 0/3            & 1/3            & 1/3            & 1/3          \\
                                                                                                 & Prompt Patterns~(6 Experiments)                 & 0/6                         & 2/6            & 0/6            & 1/6            & 3/6          \\
                                                                                                 & Tasks~(2 Experiments)                           & 0/2                         & 0/2            & 1/2            & 1/2            & 0/2          \\
                                                                                                 & \textit{Max. $wP$~score}                        & 0.4453                      & 0.4878         & 0.4103         & 0.5999         & 0.5388       \\
                                                                                                 & \textit{Avg. ~$wP$score}                        & 0.12                        & 0.12           & 0.12           & 0.13           & 0.12         \\
$wR$                                                                                             & LLMs~(3 Experiments)                            & 1/3                         & 0/3            & 1/3            & 1/3            & 0/3          \\
                                                                                                 & Prompt Patterns~(6 Experiments)                 & 2/6                         & 0/6            & 1/6            & 1/6            & 2/6          \\
                                                                                                 & Tasks~(2 Experiments)                           & 1/2                         & 0/2            & 0/2            & 1/2            & 0/2          \\
                                                                                                 & \textit{Max. $wR$~score}                        & 0.4257                      & 0.3253         & 0.3173         & 0.2932         & 0.3092       \\
                                                                                                 & \textit{Avg. ~$wR$score}                        & 0.15                        & 0.14           & 0.15           & 0.16           & 0.17         
\end{tblr}
\end{table}

Table \ref{tab:text-factor-analysis} presents the effects of different dataset variations on the other experimental factors: models, prompts, and tasks. It also highlights how often each dataset variation positively influenced these factors. Below, we provide a detailed analysis of these results.

\subsubsection{Impact of Dataset Variations on Binary Classification}

The findings derived from Table \ref{tab:text-factor-analysis} are as follows:

\begin{itemize}
    \item \textbf{Weighted Precision:} Sentence completion in the requirements text positively impacts the performance of LLMs and prompt patterns. In contrast, punctuation removal has no effect on any of the experimental factors. Labels in lowercase show no impact on any factor, while uppercase labels have a mild effect on prompt patterns. Labels with capitalized first letters, however, have an effect on all three factors.
    \item \textbf{Weighted Recall:} Capitalized first letters on labels have no effect on the experimental factors. Conversely, lowercase labels appear to influence all three factors. The other dataset variations show a more mixed effect, impacting one or two factors. 
\end{itemize}

In summary, the five dataset variations have a \textit{marginal effect} on the experimental results, as both the weighted precision and weighted recall scores remain relatively uniform across different settings. This suggests that the performance differences between the various dataset variations are minimal.

\subsubsection{Impact of Dataset Variations on Multi-Class Classification}

The findings derived from Table \ref{tab:text-factor-analysis} are as follows:

\begin{itemize}
    
    \item \textbf{Weighted Precision:} Punctuation removal in the requirements text has no impact on any of the experimental factors. In contrast, sentence completion positively influences the performance of prompt patterns. Lowercase labels affect LLMs and tasks, while uppercase labels positively impact all three factors. Labels with capitalised first letters affect LLMs and prompt patterns. 
    \item \textbf{Weighted Recall:} Punctuation removal in the requirements text positively impacts all three experimental factors. On the other hand, sentence completion has no effect on any of these factors. Lowercase labels affect LLMs and prompt patterns, while uppercase labels positively influence all three factors. Labels with capitalised first letters, however, affect only one factor.
\end{itemize}

In summary, the five dataset variations have a \textit{marginal effect} on the experimental results, as both the weighted precision and weighted recall scores remain relatively consistent across different settings. This suggests that the performance differences between the various dataset variations are minimal.

\subsubsection{Key Findings and Insights}

The key results derived from our dataset variation experiments are summarized below.

\begin{tcolorbox}[colback=green!5!white,colframe=green!75!black,title=Key Findings on Dataset Variations] 
\small 
\begin{itemize} 
\item \textbf{Requirement Text Variations:} Modifying requirement text, such as removing punctuation or adding a full stop, consistently enhanced performance in both binary and multi-class tasks. Removing punctuation resulted in slightly higher peak scores for both weighted precision and recall. However, when considering average performance, both text modifications showed almost identical effects. 

\item \textbf{Label Variations:} Variations in label text (lowercase, uppercase, capitalised) had a more significant impact, especially in binary classification. Capitalised labels performed best in weighted precision, while lowercase labels excelled in weighted recall. In multi-class tasks, uppercase and capitalised labels yielded stronger overall performance. Nevertheless, the average impact of these label variations was largely similar. 
\end{itemize} 

\end{tcolorbox}

\textit{The lack of significant performance differences across dataset variations suggests that the model is highly robust to variations in requirement text and label formatting}. In zero-shot classification, the model leverages its pre-trained semantic understanding to generalise effectively across diverse inputs. Given that modern generative LLMs are trained on large, varied datasets, they are inherently capable of handling minor text modifications—such as punctuation removal or case changes—without a substantial decline in performance. This robustness indicates that the model's ability to infer meaning and classify requirements is not highly sensitive to these types of variations.

\textit{The results imply that the model prioritises semantic content over surface-level formatting.} Whether punctuation is removed or labels are capitalised, the core meaning of the requirements and labels remains unchanged. The model's pre-trained knowledge enables it to effectively extract and process semantic information, regardless of these text variations. This supports the zero-shot paradigm, where the model's strength lies in its ability to generalise based on meaning rather than specific syntactic cues.

Regarding label text variations, the absence of significant performance differences indicates that the model treats lowercase, uppercase, and capitalised labels as semantically equivalent. This behaviour aligns with how LLMs typically handle case sensitivity, often normalising text internally to focus on meaning rather than formatting. Consequently, variations in label case do not offer additional discriminative power for the model, resulting in consistent performance across all label formats.

Although these variations do not notably affect performance, the consistent results suggest that practitioners have flexibility in formatting requirements and labels for zero-shot classification. This reduces the need for strict preprocessing rules, making the approach more adaptable to real-world applications.

We summarise the key insights and observations derived from our research on dataset variations as follows:

\begin{tcolorbox}[colback=green!5!white,colframe=green!75!black,title=Key Insights into Dataset Variations]\label{RQ1LLMBox2}
\small 
\begin{itemize} 
\item \textbf{Requirement Text Variations:} The model shows remarkable resilience to changes in punctuation and text casing, highlighting its capacity to focus on the semantic meaning of the text rather than its syntactic structure. 
\item \textbf{Label Variations:} Despite minor performance differences, the model maintains consistent performance across different label formats (i.e., lowercase, uppercase, and capitalised). This indicates that the model treats these variations as semantically equivalent, reinforcing its ability to abstract and generalise from textual content without being influenced by formatting. 
\end{itemize} 
\end{tcolorbox}

\subsection{Impact of Prompt Patterns}

\begin{table}
\centering
\caption{Impact of Prompt Patterns on LLMs, Dataset Variations and Tasks. It also highlights how frequently each prompt pattern has a positive impact on these factors.}
\label{tab:prompt-factor-analysis}
\scriptsize 
\begin{tblr}{
  row{1} = {c},
  row{2} = {c},
  row{3} = {c},
  row{14} = {c},
  row{15} = {c},
  row{16} = {c},
  cell{1}{1} = {c=8}{},
  cell{2}{1} = {r=2}{},
  cell{2}{2} = {r=2}{},
  cell{2}{3} = {c=6}{},
  cell{4}{1} = {r=5}{c},
  cell{4}{3} = {c},
  cell{4}{4} = {c},
  cell{4}{5} = {c},
  cell{4}{6} = {c},
  cell{4}{7} = {c},
  cell{4}{8} = {c},
  cell{5}{3} = {c},
  cell{5}{4} = {c},
  cell{5}{5} = {c},
  cell{5}{6} = {c},
  cell{5}{7} = {c},
  cell{5}{8} = {c},
  cell{6}{3} = {c},
  cell{6}{4} = {c},
  cell{6}{5} = {c},
  cell{6}{6} = {c},
  cell{6}{7} = {c},
  cell{6}{8} = {c},
  cell{7}{3} = {c},
  cell{7}{4} = {c},
  cell{7}{5} = {c},
  cell{7}{6} = {c},
  cell{7}{7} = {c},
  cell{7}{8} = {c},
  cell{8}{3} = {c},
  cell{8}{4} = {c},
  cell{8}{5} = {c},
  cell{8}{6} = {c},
  cell{8}{7} = {c},
  cell{8}{8} = {c},
  cell{9}{1} = {r=5}{c},
  cell{9}{3} = {c},
  cell{9}{4} = {c},
  cell{9}{5} = {c},
  cell{9}{6} = {c},
  cell{9}{7} = {c},
  cell{9}{8} = {c},
  cell{10}{3} = {c},
  cell{10}{4} = {c},
  cell{10}{5} = {c},
  cell{10}{6} = {c},
  cell{10}{7} = {c},
  cell{10}{8} = {c},
  cell{11}{3} = {c},
  cell{11}{4} = {c},
  cell{11}{5} = {c},
  cell{11}{6} = {c},
  cell{11}{7} = {c},
  cell{11}{8} = {c},
  cell{12}{3} = {c},
  cell{12}{4} = {c},
  cell{12}{5} = {c},
  cell{12}{6} = {c},
  cell{12}{7} = {c},
  cell{12}{8} = {c},
  cell{13}{3} = {c},
  cell{13}{4} = {c},
  cell{13}{5} = {c},
  cell{13}{6} = {c},
  cell{13}{7} = {c},
  cell{13}{8} = {c},
  cell{14}{1} = {c=8}{},
  cell{15}{1} = {r=2}{},
  cell{15}{2} = {r=2}{},
  cell{15}{3} = {c=6}{},
  cell{17}{1} = {r=5}{c},
  cell{17}{3} = {c},
  cell{17}{4} = {c},
  cell{17}{5} = {c},
  cell{17}{6} = {c},
  cell{17}{7} = {c},
  cell{17}{8} = {c},
  cell{18}{3} = {c},
  cell{18}{4} = {c},
  cell{18}{5} = {c},
  cell{18}{6} = {c},
  cell{18}{7} = {c},
  cell{18}{8} = {c},
  cell{19}{3} = {c},
  cell{19}{4} = {c},
  cell{19}{5} = {c},
  cell{19}{6} = {c},
  cell{19}{7} = {c},
  cell{19}{8} = {c},
  cell{20}{3} = {c},
  cell{20}{4} = {c},
  cell{20}{5} = {c},
  cell{20}{6} = {c},
  cell{20}{7} = {c},
  cell{20}{8} = {c},
  cell{21}{3} = {c},
  cell{21}{4} = {c},
  cell{21}{5} = {c},
  cell{21}{6} = {c},
  cell{21}{7} = {c},
  cell{21}{8} = {c},
  cell{22}{1} = {r=5}{c},
  cell{22}{3} = {c},
  cell{22}{4} = {c},
  cell{22}{5} = {c},
  cell{22}{6} = {c},
  cell{22}{7} = {c},
  cell{22}{8} = {c},
  cell{23}{3} = {c},
  cell{23}{4} = {c},
  cell{23}{5} = {c},
  cell{23}{6} = {c},
  cell{23}{7} = {c},
  cell{23}{8} = {c},
  cell{24}{3} = {c},
  cell{24}{4} = {c},
  cell{24}{5} = {c},
  cell{24}{6} = {c},
  cell{24}{7} = {c},
  cell{24}{8} = {c},
  cell{25}{3} = {c},
  cell{25}{4} = {c},
  cell{25}{5} = {c},
  cell{25}{6} = {c},
  cell{25}{7} = {c},
  cell{25}{8} = {c},
  cell{26}{3} = {c},
  cell{26}{4} = {c},
  cell{26}{5} = {c},
  cell{26}{6} = {c},
  cell{26}{7} = {c},
  cell{26}{8} = {c},
  hline{1-2,4,9,14-15,17,22,27} = {-}{},
  hline{3,16} = {3-8}{},
  hline{7,12,20,25} = {2-8}{dashed},
}
\textbf{A) Effects of Prompt Patterns on Other Factors in Binary Classification Setting}      &                                              &                                                     &                                                       &                                              &                                                 &                                                     &                                                       \\
\textbf{Metrics}                                                                              & \textbf{Factors affected by Prompt Patterns} & \textbf{Prompt Patterns}                            &                                                       &                                              &                                                 &                                                     &                                                       \\
                                                                                              &                                              & {\textbf{is-}\\\textbf{about}\\\textbf{~assertion}} & {\textbf{belongs-}\\\textbf{to }\\\textbf{assertion}} & {\textbf{is-}\\\textbf{about}\\\textbf{Q/A}} & {\textbf{belongs-}\\\textbf{to }\\\textbf{Q/A}} & {\textbf{is-}\\\textbf{about}\\\textbf{definition}} & {\textbf{belongs-}\\\textbf{to}\\\textbf{definition}} \\
$wP$                                                                                          & LLMs~(3 Experiments)                         & 0/3                                                 & 1/3                                                   & 0/3                                          & 0/3                                             & 0/3                                                 & 2/3                                                   \\
                                                                                              & Dataset Variations (5 Experiments)           & 0/5                                                 & 2/5                                                   & 0/5                                          & 0/5                                             & 0/5                                                 & 3/5                                                   \\
                                                                                              & Tasks~(3 Experiments)                        & 0/3                                                 & 1/3                                                   & 0/3                                          & 1/3                                             & 0/3                                                 & 1/3                                                   \\
                                                                                              & \textit{Max. $wP$ score}                     & 0.7683                                              & 0.7680                                                & 0.6883                                       & 0.7303                                          & 0.6519                                              & 0.7611                                                \\
                                                                                              & \textit{Avg. ~$wP$ score}                    & 0.46                                                & 0.56                                                  & 0.50                                         & 0.52                                            & 0.41                                                & 0.39                                                  \\
$wR$                                                                                          & LLMs~(3 Experiments)                         & 0/3                                                 & 2/3                                                   & 1/3                                          & 0/3                                             & 0/3                                                 & 0/3                                                   \\
                                                                                              & Dataset Variations (5 Experiments)           & 0/5                                                 & 0/5                                                   & 2/5                                          & 0/5                                             & 0/5                                                 & 3/5                                                   \\
                                                                                              & Tasks~(3 Experiments)                        & 0/3                                                 & 1/3                                                   & 1/3                                          & 1/3                                             & 0/3                                                 & 0/3                                                   \\
                                                                                              & \textit{Max. $wR $score}                     & 0.6157                                              & 0.6527                                                & 0.6588                                       & 0.6077                                          & 0.6266                                              & 0.6454                                                \\
                                                                                              & \textit{Avg. $wR$ score}                     & 0.47                                                & 0.53                                                  & 0.49                                         & 0.50                                            & 0.52                                                & 0.50                                                  \\
\textbf{B) Effects of Prompt Patterns on Other Factors in Multi-class Classification Setting} &                                              &                                                     &                                                       &                                              &                                                 &                                                     &                                                       \\
\textbf{Metrics }                                                                             & \textbf{Factors affected by Prompt Patterns} & \textbf{Prompt Patterns}                            &                                                       &                                              &                                                 &                                                     &                                                       \\
                                                                                              &                                              & {\textbf{is-}\\\textbf{about}\\\textbf{assertion}}  & {\textbf{belongs-}\\\textbf{to }\\\textbf{assertion}} & {\textbf{is-}\\\textbf{about}\\\textbf{Q/A}} & {\textbf{belongs-}\\\textbf{to }\\\textbf{Q/A}} & {\textbf{is-}\\\textbf{about}\\\textbf{definition}} & {\textbf{belongs-}\\\textbf{to}\\\textbf{definition}} \\
$wP$                                                                                          & LLMs~(3 Experiments)                         & 1/3                                                 & 1/3                                                   & 0/3                                          & 0/3                                             & 1/3                                                 & 0/3                                                   \\
                                                                                              & Dataset Variations (5 Experiments)           & 2/5                                                 & 1/5                                                   & 0/5                                          & 0/5                                             & 0/5                                                 & 2/5                                                   \\
                                                                                              & Tasks~(2 Experiments)                        & 2/2                                                 & 0/2                                                   & 0/2                                          & 0/2                                             & 0/2                                                 & 0/2                                                   \\
                                                                                              & \textit{Max. $wP$ score}                     & 0.5999                                              & 0.4406                                                & 0.2571                                       & 0.3368                                          & 0.4048                                              & 0.5389                                                \\
                                                                                              & \textit{Avg. $wP$ score}                     & 0.14                                                & 0.13                                                  & 0.10                                         & 0.09                                            & 0.14                                                & 0.14                                                  \\
$wR$                                                                                          & LLMs~(3 Experiments)                         & 1/3                                                 & 0/3                                                   & 1/3                                          & 0/3                                             & 1/3                                                 & 0/3                                                   \\
                                                                                              & Dataset Variations (5 Experiments)           & 3/5                                                 & 0/5                                                   & 0/5                                          & 0/5                                             & 1/5                                                 & 1/5                                                   \\
                                                                                              & Tasks~(2 Experiments)                        & 1/2                                                 & 0/2                                                   & 0/2                                          & 0/2                                             & 1/2                                                 & 0/2                                                   \\
                                                                                              & \textit{Max. $wR$ score}                     & 0.4257                                              & 0.2731                                                & 0.2932                                       & 0.2691                                          & 0.3092                                              & 0.3012                                                \\
                                                                                              & \textit{Avg. $wR$ score}                     & 0.16                                                & 0.14                                                  & 0.15                                         & 0.13                                            & 0.17                                                & 0.17                                                  
\end{tblr}
\end{table}

Table \ref{tab:prompt-factor-analysis} presents the effects of different prompt patterns on the experimental factors: models, dataset variations, and tasks. It also highlights how frequently each prompt pattern positively influenced these factors. Below, we provide a detailed analysis of these results.

\subsubsection{Impact of Prompt Patterns on Binary Classification}

The findings derived from Table \ref{tab:prompt-factor-analysis} are as follows:

\begin{itemize}
    \item \textbf{Weighted Precision:} A striking observation is that all three ``is-about'' prompt patterns—``is-about assertion,'' ``is-about Q/A,'' and ``is-about definition''—had no effect on any of the experimental factors, whereas all three ``belongs-to'' patterns exhibited a varied degree of impact on these factors. Despite this, while the ``is-about assertion'' pattern had no effect on any of the experimental factors, it achieved the highest precision score ($Max.wP=0.7683$). Although the ``belongs-to assertion'' pattern came second in terms of precision, it achieved the highest average precision score ($Avg.wP=0.56$). These results indicate that the ``belongs-to assertion'' prompt pattern is the best performer, as measured by weighted precision.
    
    \item \textbf{Weighted Recall:} Two of the three ``is-about'' prompt patterns—``is-about assertion'' and ``is-about definition''—had no effect on any of the experimental factors, but still achieved solid performance comparable to the other prompt patterns. Both the ``is-about Q/A'' and ``belongs-to assertion'' patterns delivered robust performance, with the former achieving the highest recall score ($Max.wR=0.6588$) and the latter the highest average recall ($Avg.wR=0.53$). 
\end{itemize}

In summary, the ``is-about assertion'' prompt pattern has the most influence on binary classification tasks and is thus the best-performing pattern. The remaining five prompt patterns have a \textit{similar effect} on binary classification performance, as both the weighted precision and weighted recall scores remain relatively consistent across different settings. This suggests that the performance differences among the remaining five prompt patterns are minimal. 

\subsubsection{Impact of Prompt Patterns on Multi-Class Classification}

The findings derived from Table \ref{tab:prompt-factor-analysis} are as follows:

\begin{itemize}
    \item \textbf{Weighted Precision:} An interesting observation is that the two Q/A prompt patterns---``is-about Q/A'' and ``belongs-to Q/A,''---had no effect on any of the experimental factors. Furthermore, these two patterns were the worst performers compared to the others, as indicated by their poor precision scores. The ``is-about assertion'' pattern is the best performer, achieving the highest precision score ($Max.wP=0.5999$) and matching the ``is-about definition'' and ``belongs-to definition'' patterns in terms of average precision ($Avg.wP=0.14$).
    
    \item \textbf{Weighted Recall:} Two of the three ``belongs-to'' prompt patterns—``belongs-to assertion'' and ``belongs-to Q/A''—had no effect on any of the experimental factors and were also the worst performers, as shown by their low recall scores. In contrast, the ``is-about assertion'' pattern is the best performer, achieving the highest recall score ($Max.wR=0.4257$). Its average recall score ($Avg.wR=0.16$) is only slightly lower than the highest average recall score ($Avg.wR=0.17$).
\end{itemize}

In summary, the ``is-about assertion'' prompt pattern has the most influence on multi-class classification tasks, making it the best-performing pattern. By contrast, the Q/A prompt patterns had no effect on any of the experimental factors and were the worst performers.

\subsubsection{Key Findings and Insights} 

The key results derived from our prompt pattern experiments are summarised below.

\begin{tcolorbox}[colback=green!5!white,colframe=green!75!black,title=Key Findings on Prompt Patterns]
\small
\begin{itemize}
    \item \textbf{Assertion-Based Patterns} consistently perform well across both binary and multi-class classifications, achieving the highest maximum scores and competitive average scores.
    \item \textbf{Q/A-Based Patterns} are effective for binary classification, particularly in $wR$ measures, but perform poorly in multi-class classification.
    \item \textbf{Definition-Based Patterns} exhibit varying performance, showing strong results in multi-class classification (especially in $wR$) but less consistency in binary classification.
\end{itemize}
\end{tcolorbox}

An important finding from this study is that \textit{the choice of prompt pattern can significantly impact the outcomes of requirement classification tasks}. This is evident in the variability of how frequently each prompt achieved the best performance, as well as the differences observed in the average and maximum scores when evaluating the weighted metrics $wP$ and $wR$. Further analysis reveals that some patterns excel in binary classification tasks, while others perform better in multi-class classification scenarios. This suggests that the effectiveness of prompt patterns is highly \textit{task-dependent} and should be tailored to the specific nature of the classification task.

For example, assertion-based patterns demonstrate consistent performance across both binary and multi-class tasks due to their simplicity and directness, making them a reliable choice for general use. In contrast, Q/A-based patterns are particularly well suited for binary classification, where the yes/no or multiple-choice format aligns naturally with the task structure. However, their rigid format limits their applicability in multi-class scenarios, where the diversity and complexity of labels require a more flexible approach.

Similarly, definition-based patterns perform strongly in multi-class classification, as their detailed explanations help the model distinguish between closely related labels. However, their added complexity is often unnecessary for binary classification, where simpler patterns are more effective. This variability highlights the importance of selecting the appropriate prompt pattern based on the task's specific requirements and the nature of the labels involved.

Ultimately, while no single prompt pattern is universally superior, understanding their strengths and limitations enables more informed decisions in designing effective classification tasks.

We summarise the key insights and observations derived from our research on prompt patterns as follows:

\begin{tcolorbox}[colback=green!5!white,colframe=green!75!black,title=Key Insights into Prompt Patterns]
\small
  \begin{itemize}
    \item \textbf{Assertion-Based Patterns} are simple and direct, making them highly effective for classification tasks. By clearly stating a fact or claim, they align well with both binary and multi-class scenarios. Their straightforward nature minimises ambiguity, leading to high accuracy and consistency across various classification types.

    \item \textbf{Q/A-Based Patterns} rely on yes/no or multiple-choice questions, which are well-suited for binary classification (e.g., ``Is this correct? Yes/No''). However, in multi-class classification, where labels are more complex and diverse, this approach becomes less effective. The rigid yes/no or multiple-choice format struggles to distinguish between closely related or semantically similar labels.

    \item \textbf{Definition-Based Patterns} provide detailed explanations or descriptions of labels, making them particularly useful in multi-class classification. This approach helps the model discern subtle differences between closely related labels, enhancing performance in complex scenarios. However, in binary classification, where the task is simpler (e.g., two options), the added complexity may be unnecessary, leading to less consistent results.
\end{itemize}
\end{tcolorbox}
\subsection{Impact of Task Types}

\begin{table}
\centering
\caption{Impact of LLMs, Dataset Variations and Prompt Patterns on Task Performance. It also highlights how frequently each task has been affected by these factors.}
\label{tab:task-factor-analysis}
\scriptsize
\begin{tblr}{
  row{1} = {c},
  row{2} = {c},
  row{3} = {c},
  row{14} = {c},
  row{15} = {c},
  row{16} = {c},
  cell{1}{1} = {c=6}{},
  cell{2}{1} = {r=2}{},
  cell{2}{2} = {r=2}{},
  cell{2}{3} = {c=4}{},
  cell{3}{4} = {c=2}{},
  cell{4}{1} = {r=5}{c},
  cell{4}{3} = {c},
  cell{4}{4} = {c=2}{c},
  cell{4}{6} = {c},
  cell{5}{3} = {c},
  cell{5}{4} = {c=2}{c},
  cell{5}{6} = {c},
  cell{6}{3} = {c},
  cell{6}{4} = {c=2}{c},
  cell{6}{6} = {c},
  cell{7}{3} = {c},
  cell{7}{4} = {c=2}{c},
  cell{7}{6} = {c},
  cell{8}{3} = {c},
  cell{8}{4} = {c=2}{c},
  cell{8}{6} = {c},
  cell{9}{1} = {r=5}{c},
  cell{9}{3} = {c},
  cell{9}{4} = {c=2}{c},
  cell{9}{6} = {c},
  cell{10}{3} = {c},
  cell{10}{4} = {c=2}{c},
  cell{10}{6} = {c},
  cell{11}{3} = {c},
  cell{11}{4} = {c=2}{c},
  cell{11}{6} = {c},
  cell{12}{3} = {c},
  cell{12}{4} = {c=2}{c},
  cell{12}{6} = {c},
  cell{13}{3} = {c},
  cell{13}{4} = {c=2}{c},
  cell{13}{6} = {c},
  cell{14}{1} = {c=6}{},
  cell{15}{1} = {r=2}{},
  cell{15}{2} = {r=2}{},
  cell{15}{3} = {c=4}{},
  cell{16}{3} = {c=2}{},
  cell{16}{5} = {c=2}{},
  cell{17}{1} = {r=5}{c},
  cell{17}{3} = {c=2}{c},
  cell{17}{5} = {c=2}{c},
  cell{18}{3} = {c=2}{c},
  cell{18}{5} = {c=2}{c},
  cell{19}{3} = {c=2}{c},
  cell{19}{5} = {c=2}{c},
  cell{20}{3} = {c=2}{c},
  cell{20}{5} = {c=2}{c},
  cell{21}{3} = {c=2}{c},
  cell{21}{5} = {c=2}{c},
  cell{22}{1} = {r=5}{c},
  cell{22}{3} = {c=2}{c},
  cell{22}{5} = {c=2}{c},
  cell{23}{3} = {c=2}{c},
  cell{23}{5} = {c=2}{c},
  cell{24}{3} = {c=2}{c},
  cell{24}{5} = {c=2}{c},
  cell{25}{3} = {c=2}{c},
  cell{25}{5} = {c=2}{c},
  cell{26}{3} = {c=2}{c},
  cell{26}{5} = {c=2}{c},
  hline{1-2,4,9,14-15,17,22,27} = {-}{},
  hline{3,16} = {3-6}{},
  hline{7,12,20,25} = {2-6}{dashed},
}
\textbf{A) Effects on Binary Classification Tasks by Other Factors}      &                                    &                          &                       &                        &                        \\
\textbf{Metrics }                                                        & \textbf{Factors affecting Tasks}   & \textbf{Tasks}           &                       &                        &                        \\
                                                                         &                                    & \textbf{Task Functional} & \textbf{Task Quality} &                        & \textbf{Task Security} \\
$wP$                                                                     & LLMs~(3 Experiments)               & 2/3                      & 0/3                   &                        & 1/3                    \\
                                                                         & Dataset Variations~(5 Experiments) & 2/5                      & 0/5                   &                        & 3/5                    \\
                                                                         & Prompt Patterns~(6 Experiments)    & 1/6                      & 3/6                   &                        & 2/6                    \\
                                                                         & \textit{Max. $wP$ score}           & 0.7611                   & 0.7303                &                        & 0.7683                 \\
                                                                         & \textit{Avg. ~$wP$ score}          & 0.46                     & 0.51                  &                        & 0.45                   \\
$wR$                                                                     & LLMs~(3 Experiments)               & 2/3                      & 0/3                   &                        & 1/3                    \\
                                                                         & Dataset Variations~(5 Experiments) & 3/5                      & 0/5                   &                        & 2/5                    \\
                                                                         & Prompt patterns~(6 Experiments)    & 3/6                      & 1/6                   &                        & 2/6                    \\
                                                                         & \textit{Max. $wR$ score}           & 0.6527                   & 0.6077                &                        & 0.6588                 \\
                                                                         & \textit{Avg. ~$wR$ score}          & 0.53                     & 0.51                  &                        & 0.47                   \\
\textbf{B) Effects on Multi-class Classification Tasks by Other Factors} &                                    &                          &                       &                        &                        \\
\textbf{Metrics }                                                        & \textbf{Factors affecting Tasks}   & \textbf{Tasks}           &                       &                        &                        \\
                                                                         &                                    & \textbf{Task NFR}        &                       & \textbf{Task NFR-Top4} &                        \\
$wP$                                                                     & LLMs~(3 Experiments)               & 0/3                      &                       & 3/3                    &                        \\
                                                                         & Dataset Variations~(5 Experiments) & 0/5                      &                       & 5/5                    &                        \\
                                                                         & Prompt Patterns~(6 Experiments)    & 0/6                      &                       & 6/6                    &                        \\
                                                                         & \textit{Max. $wP$ score}           & 0.2694                   &                       & 0.5999                 &                        \\
                                                                         & \textit{Avg. ~$wP$ score}          & 0.05                     &                       & 0.19                   &                        \\
$wR$                                                                     & LLMs~(3 Experiments)               & 0/3                      &                       & 3/3                    &                        \\
                                                                         & Dataset Variations~(5 Experiments) & 0/5                      &                       & 5/5                    &                        \\
                                                                         & Prompt Patterns~(6 Experiments)    & 0/6                      &                       & 6/6                    &                        \\
                                                                         & \textit{Max. $wR$ score}           & 0.2493                   &                       & 0.4257                 &                        \\
                                                                         & \textit{Avg. ~$wR$ score}          & 0.08                     &                       & 0.22                   &                        
\end{tblr}
\end{table}

Table \ref{tab:task-factor-analysis} presents the effects of different experimental factors on tasks. It also highlights the frequency with which each task is affected by these factors. Below, we provide a detailed analysis of these results.

\subsubsection{Impact of Binary Classification Tasks}

The findings derived from Table \ref{tab:task-factor-analysis} are as follows:

\begin{itemize}
    \item \textbf{Weighted Precision:} Task Security leads, being most positively impacted by the other experimental factors. It achieved the highest maximum precision score ($Max.wP=0.7683$), though its average precision score is the lowest ($Avg.wP=0.45$). In contrast, Task Quality achieved the highest average precision score ($Avg.wP=0.51$). Task Functional performed moderately, with a maximum precision score of $Max.wP=0.7611$ and an average precision score of $Avg.wP=0.46$.

    \item \textbf{Weighted Recall:} Task Security also achieved the highest maximum recall score ($Max.wR=0.6588$), but its average recall score was the lowest ($Avg.wR=0.47$). In contrast, Task Functional had the highest average recall score ($Avg.wR=0.53$). Task Quality performed moderately, achieving a maximum recall score of $Max.wR=0.6527$. Interestingly, Task Quality exhibited consistent performance across different settings, with identical average precision and recall scores of $0.51$.
\end{itemize}

In summary, both Task Functional and Task Security have been positively affected by different LLMs, dataset variations and prompt patterns. These two tasks are also easier to perform than Task Quality. In contrast, Task Quality is a more difficult task, but it exhbits the consistency in its average precision and recall scores.In summary, both Task Functional and Task Security were positively affected by different LLMs, dataset variations, and prompt patterns. These tasks were easier to perform compared to Task Quality. Task Quality, although more challenging, showed consistent average precision and recall scores.

\subsubsection{Impact of Multi-Class Classification Tasks}

The findings derived from Table \ref{tab:task-factor-analysis} are as follows:

\begin{itemize}
    \item \textbf{Task NFR:} None of the three experimental factors impacted the performance of Task NFR. This task performed poorly under both weighted precision and weighted recall, achieving an average weighted precision score of $0.05$ and an average weighted recall of $0.08$.

    \item \textbf{Task NFR-Top4:} In contrast, all three experimental factors positively impacted the performance of Task NFR-Top4. This task showed solid performance in both weighted precision and weighted recall, with an average weighted precision score of $0.19$ and an average weighted recall score of $0.22$.
\end{itemize}

In summary, Task NFR proved to be challenging, with very poor performance. Conversely, Task NFR-Top4 showed more satisfactory performance.

\subsubsection{Key Findings and Insights} 

The key results derived from our task experiments are summarised below.

\begin{tcolorbox}[colback=green!5!white,colframe=green!75!black,title=Key Findings on Tasks]
\small

\begin{itemize}

    \item \textbf{Binary classification tasks} generally outperformed multi-class tasks, achieving higher maximum and average scores. \textbf{Task Functional} performed steadily, showing strong consistency and frequently ranking as the best performer. \textbf{Task Security} stood out with the highest peak performance, while Quality, though moderate overall, achieved the highest average performance in the weighted precision measure.

    \item \textbf{Multi-class classification Task NFR} performed poorly, with very low scores in both weighted precision and recall, indicating significant challenges across all metrics. In contrast, \textbf{Task NFR-Top4} emerged as the clear leader, achieving the highest scores in both performance measures and consistently outperforming Task NFR.
 
\end{itemize}
\end{tcolorbox}

These results reveal distinct performance trends between binary and multi-class classification tasks, driven by both task structure and label semantics. In binary classification, tasks such as \textbf{Task Functional, Task Security,} and \textbf{Task Quality} showed strong and consistent performance. \textbf{Task Security} achieved the highest peak scores, likely due to its clear semantic focus on protection and risk mitigation, which simplifies classification. \textbf{Task Functional} demonstrated the most consistency, as its semantic focus on system behaviour and features aligns well with binary distinctions. \textbf{Task Quality}, while moderate overall, excelled in average precision, benefiting from its broad yet well-defined semantic scope related to system performance and usability.

In multi-class classification, the semantic complexity of labels had a more significant impact on performance. \textbf{Task NFR} underperformed substantially, likely due to its broad and often ambiguous semantic scope, as well as the large number of classes—10 in total. The classes varied significantly in size, with the largest having 67 instances and the smallest only 10. This diversity in class size, combined with semantic ambiguity, posed challenges for accurate classification. In contrast, \textbf{Task NFR-Top4} focused only on the four largest classes, which were more balanced in size and had clearer, well-defined semantics (e.g., performance, security, usability, and operational). This focus on specific, clearly defined categories helped the model achieve better performance.

These findings underscore \textit{the significant role of label semantics in classification performance}. Binary tasks benefit from clear, distinct labels, while multi-class tasks require narrower, semantically well-defined categories to optimise results. This highlights the importance of considering label semantics when designing classification systems.

\begin{tcolorbox}[colback=green!5!white,colframe=green!75!black,title=Key Insights into Tasks]
\small
    \begin{itemize}
    \item \textbf{Binary classification tasks} generally perform better due to their simpler two-category structure. Tasks like Security and Functional excel because of their clear and distinct semantic focus, while Quality demonstrates strong average performance despite being more moderate overall.

    \item \textbf{Multi-class classification tasks} are more challenging due to the complexity of managing multiple, often complex labels. NFR struggles due to its broad and ambiguous semantic scope, whereas NFR-Top4 succeeds by focusing on a narrower, well-defined subset of labels, resulting in higher precision and recall.
    \end{itemize}
\end{tcolorbox}

\subsection{Statistical Signifcance Analysis}

\begin{table}
\centering
\caption{Friedman test results for each of four factors (i.e., LLMs, prompt patterns, dataset variations, and tasks) with potential impact to binary and multiclass classification results. The test results are calculated based on $wP$, $wR$, and $wF1$ performance results, independently. Cells marked by (*) refer to a rejection of the null hypothesis when $p$-value $< 0.05$. }
\label{tab:stat-friedman}
\scriptsize
\begin{tblr}{
  cells = {c},
  cell{1}{1} = {c=5}{},
  cell{6}{1} = {c=5}{},
  hline{1-3,6-8,11} = {-}{},
}
\textbf{A) Binary Classification Results}      &                                              &                                                         &                                                            &                                               \\
\textbf{Measure}                               & {\textbf{LLMs}\\\textbf{$p$-value}}          & {\textbf{Prompt Patterns}\\\textbf{\textbf{$p$-value}}} & {\textbf{Dataset Variations}\\\textbf{\textbf{$p$-value}}} & {\textbf{Tasks}\\\textbf{\textbf{$p$-value}}} \\
\textbf{$wP$}                                  & 0.02355 *                                    & 0.00012 *                                               & 0.69655                                                    & 0.19311                                       \\
\textbf{$wR$}                                  & 0.45469                                      & 0.00016 *                                               & 0.95959                                                    & 0.00037 *                                     \\
\textbf{$wF1$}                                 & 0.00757 *                                    & 0.00167 *                                               & 0.88053                                                    & 0.00110 *                                     \\
\textbf{B) Multi-class Classification Results} &                                              &                                                         &                                                            &                                               \\
\textbf{Measure}                               & {\textbf{LLMs}\\\textbf{\textbf{$p$-value}}} & {\textbf{Prompt Patterns}\\\textbf{\textbf{$p$-value}}} & {\textbf{Datasets}\\\textbf{\textbf{$p$-value}}}           & {\textbf{Tasks}\\\textbf{\textbf{$p$-value}}} \\
\textbf{$wP$}                                  & 1.229e-09 *                                  & 0.67237                                                 & 0.50031                                                    & 5.894e-22 *                                   \\
\textbf{$wR$}                                  & 7.112e-15 *                                  & 9.009e-05 *                                             & 0.00331 *                                                  & 1.994e-25 *                                   \\
\textbf{$wF1$}                                 & 0.00020 *                                    & 0.00162 *                                               & 0.42754                                                    & 1.694e-28 *                                   
\end{tblr}
\end{table}

The insights drawn from the Friedman Test results in Table~\ref{tab:stat-friedman} highlight the influence of different factors on model performance in classification tasks. In binary classification, the significant impact of prompt patterns on all metrics ($wP$, $wR$, $wF1$) emphasises the critical role of how information is structured and presented to the LLM. This suggests that prompt design and formulation are more impactful than previously thought, directly influencing the model's ability to process and classify requirements accurately.

Interestingly, dataset variations do not show a significant impact in binary classification, which may imply that once a model is robustly trained, minor variations in the dataset do not substantially affect its performance. This resilience could be beneficial in real-world applications where data inconsistencies are common, suggesting that trained models can maintain high performance despite such variations.

To further justify the $p$-values presented in Table \ref{tab:stat-friedman}, we examined the average scores in the summary performance tables (Tables \ref{tab:LLM-factor-analysis}, \ref{tab:prompt-factor-analysis}, \ref{tab:text-factor-analysis}, and \ref{tab:task-factor-analysis}). We looked for notable differences in the average scores for each measure (e.g., $wP$ and $wR$) for each target factor, as such differences can indicate the influence of the factor on classification performance.

For instance, when analysing the dataset variation factor (Table \ref{tab:text-factor-analysis}), we observed that the overall average scores for $wP$ and $wR$ were nearly identical ($\pm 0.01$), except for $wR$ in multi-class classification. This closeness in mean scores helps explain why most alternative hypotheses were rejected for this factor, with the exception of $wR$ in multi-class classification. Here, slight variations in average performance led to significant results (i.e., when $p$-value = $0.00331 < 0.05$), as indicated in Table \ref{tab:stat-friedman}.

In contrast, in multi-class classification, both LLMs and tasks had a strong influence across all performance measures. This underscores the importance of the model architecture and the specific nature of the tasks it is designed to handle. The results suggest that aligning a model's underlying structure with the requirements of specific tasks is central for achieving high accuracy and efficiency in more complex classification scenarios.

Overall, the results from the Friedman Test suggest that while factors like prompt design and LLM architecture are universally important, others—such as dataset variations—have a more situational impact, depending on the complexity of the classification task. This insight can guide future model development and deployment strategies, focusing on optimising prompt structures and aligning model architectures with task-specific needs for improved performance.

\subsection{Addressing Research Question RQ1}

\subsubsection{Impact of Generative LLMs on Requirements Classification}

\begin{itemize}
    \item \textbf{Binary Classification:} The results suggest that the impact of LLMs varies, with a significant influence observed only on the weighted F1-score. This indicates that while the choice of LLM does not consistently affect precision or recall, it plays a central role in balancing these metrics to optimise overall performance.
    \item \textbf{Multi-class Classification:} LLMs have a significant impact across all weighted metrics, underscoring their critical role in more complex classification scenarios. This suggests that the LLM's architecture and capabilities are vital for handling the intricacies and diversity of multi-class categorisation.
\end{itemize}

\subsubsection{Impact of Dataset Variations on Requirements Classification}

\begin{itemize}
    \item \textbf{Binary Classification:} Dataset variations do not significantly affect any of the performance metrics, indicating that the model is robust to changes in the dataset in simpler classification tasks.
    \item \textbf{Multi-class Classification:} While dataset variations significantly impact recall, they do not show a notable effect on other metrics. This suggests that the complexity of the dataset influences the model's ability to correctly identify all relevant categories but does not necessarily alter the precision or overall effectiveness of the predictions.
\end{itemize}

\subsubsection{Impact of Prompt Patterns on Requirements Classification}

\begin{itemize}
    \item \textbf{Binary Classification:} Prompt patterns significantly influence all performance metrics, highlighting that how information is presented to the model-—through the structure and wording of prompts-—is necessary for guiding the model's understanding and processing of requirements.
    \item \textbf{Multi-class Classification:} Prompt patterns have a significant impact on recall and F1-score, indicating that while prompt design may not always affect precision directly, it is vital for enhancing the model's ability to correctly identify relevant classes and strike a balance between precision and recall.
\end{itemize}

\subsubsection{Impact of Tasks on Requirements Classification}

\begin{itemize}
    \item \textbf{Binary Classification:} Tasks significantly influence weighted recall and F1-score, but not precision. This suggests that the task type affects how the model identifies relevant information and synthesises it into accurate classifications, particularly when balancing sensitivity and specificity.
    \item \textbf{Multi-class Classification:} Tasks significantly affect all metrics, emphasising the importance of task design in complex classification scenarios. The nature of the task is critical in determining how well the model can differentiate between varied and potentially overlapping classes.
\end{itemize}

\section{Experimental Results for RQ2}
\label{sec:res4RQ2}

In this section, we present and analyze the experimental results for the second research question (RQ2): \textit{How do generative LLMs perform in requirements classification compared to non-generative LLMs?} We begin by comparing the performance of the top-performing generative and non-generative LLMs on binary classification tasks, followed by an evaluation of their performance on multi-class classification tasks. We then examine the performance differences between the two types of models using the Wilcoxon Signed-Rank Test. Finally, based on these findings, we provide a comprehensive answer to the research question.

\subsection{Comparison of Top-Performing Generative and Non-Generative LLMs on Binary Classification Tasks}

Among the three binary classification tasks, our experimental results for RQ1 reveal that Llama with the ``is-about definition'' prompt performs best for \textbf{Task Functional}, while Gemma with the ``is-about Q/A'' prompt excels in \textbf{Task Quality}. Additionally, Gemma with the ``is-about Q/A'' prompt outperforms the others in \textbf{Task Security}. These results indicate that no single generative LLM performs best across all three binary classification tasks.

In this section, we compare two generative LLMs—Llama and Gemma—with two baseline non-generative models SBERT and All-Mini. Table \ref{tab:binary-best-results} displays the performance results of these four models across individual classes, measured by precision, recall, and F1 scores. It also provides the weighted average performance across all classes. Below, we evaluate the performance of these LLMs.

\definecolor{Shark}{rgb}{0.109,0.109,0.121}
\begin{table}
\centering
\caption{Comparison of Top-Performing Generative and Non-Generative LLMs on Binary Classification Tasks.}
\label{tab:binary-best-results}
\scriptsize 

\begin{tblr}{
  row{3} = {c},
  row{7} = {c},
  row{11} = {c},
  cell{1}{3} = {c=9}{c},
  cell{2}{1} = {c=2,r=2}{},
  cell{2}{3} = {c=3}{c},
  cell{2}{6} = {c=3}{c},
  cell{2}{9} = {c=3}{c},
  cell{4}{1} = {c},
  cell{4}{3} = {c},
  cell{4}{4} = {c},
  cell{4}{5} = {c},
  cell{4}{6} = {c},
  cell{4}{7} = {c},
  cell{4}{8} = {c},
  cell{5}{1} = {c},
  cell{5}{3} = {c},
  cell{5}{4} = {c},
  cell{5}{5} = {c},
  cell{5}{6} = {c},
  cell{5}{7} = {c},
  cell{5}{8} = {c},
  cell{6}{1} = {c=2,r=2}{},
  cell{6}{3} = {c=3}{c},
  cell{6}{6} = {c=3}{c},
  cell{6}{9} = {c=3}{c},
  cell{8}{1} = {c},
  cell{8}{3} = {c},
  cell{8}{4} = {c,fg=Shark},
  cell{8}{5} = {c,fg=Shark},
  cell{8}{6} = {c},
  cell{8}{7} = {c,fg=Shark},
  cell{8}{8} = {c,fg=Shark},
  cell{9}{1} = {c},
  cell{9}{3} = {c},
  cell{9}{4} = {c},
  cell{9}{5} = {c},
  cell{9}{6} = {c},
  cell{9}{7} = {c},
  cell{9}{8} = {c},
  cell{10}{1} = {c=2,r=2}{},
  cell{10}{3} = {c=3}{c},
  cell{10}{6} = {c=3}{c},
  cell{10}{9} = {c=3}{c},
  cell{12}{1} = {c},
  cell{12}{3} = {c},
  cell{12}{4} = {c},
  cell{12}{5} = {c},
  cell{12}{6} = {c},
  cell{12}{7} = {c},
  cell{12}{8} = {c},
  cell{13}{1} = {c},
  cell{13}{3} = {c},
  cell{13}{4} = {c},
  cell{13}{5} = {c},
  cell{13}{6} = {c},
  cell{13}{7} = {c},
  cell{13}{8} = {c},
  hline{1-2,6,10,14} = {-}{},
  hline{3-4,7-8,11-12} = {3-11}{},
}

{\textbf{Generative /}\\\textbf{Non-Generative}} & {\textbf{Prompt /}\\\textbf{Label}} & \textbf{Binary Classification Performance on Two Classes} 
&              &               &                                 &              &               &                             &             &              \\
\textbf{\textit{Task Functional}}               &                                            & \textbf{ FRs (578 Requirements) }            &              &               & \textbf{ NFRs (378 Requirements) } &              &               & \textbf{Weighted Average~}          &             &              \\
                                  &                                            & \textbf{ P }                           & \textbf{ R } & \textbf{ F1 } & \textbf{ P }                    & \textbf{ R } & \textbf{ F1 } & \textbf{wP}                 & \textbf{wR} & \textbf{wF1} \\
Llama                             & Is-about definition                        & 0.6956                                 & 0.6799       & 0.6877        & 0.5269                          & 0.5450       & 0.5358        & 0.6266                      & 0.6289      & 0.6276       \\
SBERT                             & Expert curated                             & 0.6218                                 & 0.9273       & 0.7444        & 0.5532                          & 0.1376       & 0.2203        & 0.6151                      & 0.5947      & 0.5372       \\
\textbf{\textit{Task Quality}}                  &                                            & \textbf{ Quality (522 Requirements) }               &              &               & \textbf{Non-Quality (434 Requirements) }    &              &               & \textbf{Weighted Average~} &             &              \\
                                  &                                            & \textbf{ P }                           & \textbf{ R } & \textbf{ F1 } & \textbf{ P }                    & \textbf{ R } & \textbf{ F1 } & \textbf{wP}                 & \textbf{wR} & \textbf{wF1} \\
Gemma                             & Belongs-to Q/A                               & 0.6315                                 & 0.6762       & 0.6531        & 0.5743                          & 0.5253       & 0.5487        & 0.6077                      & 0.6055      & 0.6057       \\
All-Mini                          & Expert curated                            & 0.5516                                 & 0.5019       & 0.5256        & 0.4595                          & 0.5092       & 0.4831        & 0.5052                      & 0.5098      & 0.5063       \\
\textbf{\textit{Task Security}}        &                                            & \textbf{Security (323 Requirements) }              &              &               & \textbf{ Non-Security (187 Requirements) }   &              &               & \textbf{Weighted Average~} &             &              \\
                                  &                                            & \textbf{ P }                           & \textbf{ R } & \textbf{ F1 } & \textbf{P}                      & \textbf{ R } & \textbf{ F1 } & \textbf{wP}                 & \textbf{wR} & \textbf{wF1} \\
Gemma                             & Is-about Q/A                                & 0.5916                                 & 0.2246       & 0.3256        & 0.6697                          & 0.9102       & 0.7716        & 0.6589                      & 0.6411      & 0.6081       \\
All-Mini                          & Expert curated                             & 0.5164                                 & 0.6377       & 0.5847        & 0.7707                          & 0.6738       & 0.6961        & 0.6490                      & 0.6774      & 0.6552       
\end{tblr}
\end{table}

\begin{itemize}
    \item \textbf{Task Functional:} The objective of this task is to classify each requirement into either the FR or NFR class. In this task, we compare the performance of Llama with that of SBERT. As shown in Table \ref{tab:binary-best-results}, Llama emerges as the most consistent model, achieving similar results across both the FR and NFR classes. In contrast, SBERT demonstrates inconsistency, performing well on the FR class but poorly on the NFR class. Notably, SBERT excels in recall for the FR class but struggles with recall for the NFR class. Overall, Llama outperforms SBERT on this task.
    
    \item \textbf{Task Quality:} The objective of this task is to classify each requirement into either the Quality or Non-Quality class. In this task, we compare the performance of Gemma with that of All-Mini. As shown in Table \ref{tab:binary-best-results}, Gemma stands out as the most consistent model, delivering similar results across both the Quality and Non-Quality classes. In contrast, All-Mini performs worse than Gemma, particularly in the Quality class. Overall, Gemma outperforms All-Mini on this task.
    
    \item \textbf{Task Security:} The objective of this task is to classify each requirement into either the Security or Non-Security class. In this task, we compare the performance of Gemma with that of All-Mini. As shown in Table \ref{tab:binary-best-results}, All-Mini emerges as the most consistent model, achieving similar results across both the Security and Non-Security classes. In contrast, Gemma demonstrates inconsistency, performing very well on the Non-Security class but poorly on the Security class. Notably, Gemma's recall performance on the Security class is particularly weak, while it excels in recall for the Non-Security class. Overall, All-Mini outperforms Gemma on this task.
\end{itemize}

In summary, the evaluation above shows that there is \textit{no clear winner among the compared generative and non-generative LLMs}. While the generative LLMs, Llama and Gemma, outperform their non-generative counterparts on Task Functional and Task Quality, respectively, the non-generative model All-Mini outperforms the generative LLM Gemma. Additionally, \textit{no single prompt pattern proves to be the best}, as the \textit{``is-about definition} prompt works well with Llama on Task Functional, while the \textit{``belongs-to Q/A''} prompt performs well with Gemma on Task Quality.

\subsection{Comparison of Top-Performing Generative and Non-Generative LLMs on Multi-Class Classification Tasks}

Our experimental results for RQ1 demonstrate that Llama outperforms other generative LLMs on both multi-class classification tasks. Specifically, Llama with the ``is-about definition'' prompt yields the best performance on \textbf{Task NFR}, while Llama with the ``is-about assertion'' prompt excels in \textbf{Task NFR-Top4}. In contrast, All-Mini outperforms SBERT on both multi-class classification tasks.

Since only one generative model (Llama) and one non-generative model (All-Mini) perform best on the multi-class classification tasks, in this section we compare the performance of these two models. Table \ref{tab:multi-best-results} presents the overall performance of Llama and All-Mini across all classes, measured by precision, recall, and F1 scores, along with the weighted average performance. Table \ref{tab:Top4-results} displays the performance results of these models on the individual top 4 classes, also measured by precision, recall, and F1 scores, and includes the weighted average performance across all classes. Below, we evaluate the comparative performance of these two LLMs.

\begin{table}
\centering
\caption{Comparison of Top-Performing Generative and Non-Generative LLMs on Multi-Class Classification Tasks. }
\label{tab:multi-best-results}
\scriptsize 

\begin{tblr}{
  row{7} = {c},
  cell{1}{3} = {c=6}{c},
  cell{2}{1} = {r=2}{},
  cell{2}{3} = {c=3}{c},
  cell{2}{6} = {c=3}{c},
  cell{3}{3} = {c},
  cell{3}{4} = {c},
  cell{3}{5} = {c},
  cell{3}{6} = {c},
  cell{3}{7} = {c},
  cell{3}{8} = {c},
  cell{4}{3} = {c},
  cell{4}{4} = {c},
  cell{4}{5} = {c},
  cell{4}{6} = {c},
  cell{4}{7} = {c},
  cell{4}{8} = {c},
  cell{5}{3} = {c},
  cell{5}{4} = {c},
  cell{5}{5} = {c},
  cell{5}{6} = {c},
  cell{5}{7} = {c},
  cell{5}{8} = {c},
  cell{6}{1} = {r=2}{c},
  cell{6}{3} = {c=3}{c},
  cell{6}{6} = {c=3}{c},
  cell{8}{3} = {c},
  cell{8}{4} = {c},
  cell{8}{5} = {c},
  cell{8}{6} = {c},
  cell{8}{7} = {c},
  cell{8}{8} = {c},
  cell{9}{3} = {c},
  cell{9}{4} = {c},
  cell{9}{5} = {c},
  cell{9}{6} = {c},
  cell{9}{7} = {c},
  cell{9}{8} = {c},
  hline{1,10} = {-}{0.08em},
  hline{2,6} = {-}{},
  hline{3-4,7-8} = {3-8}{},
}

{\textbf{Generative /}\\\textbf{Non-Generative}} & {\textbf{Prompt /}\\\textbf{Label}} & \textbf{Multi-Class Classification Results} 
&              &               &                             &               &                
\\
\textbf{\textit{Task NFR}}    &                                                             & \textbf{Performance across 10 NFR classes (369 Requirements)}               &              &               & \textbf{Weighted Average~}          &               &                \\
                                  &                                                             & \textbf{ P }                                & \textbf{ R } & \textbf{ F1 } & \textbf{ wP }               & \textbf{ wR } & \textbf{ wF1 } \\
Llama                             & Is-about definition                                        & 0.1498                                      & 0.1946       & 0.1585        & 0.2056                      & 0.2493        & 0.2094         \\
All-Mini                          & Expert curated                                              & 0.4078                                      & 0.4058       & 0.3347        & 0.5461                      & 0.3654        & 0.3763         \\
\textbf{\textit{Task NFR-Top4}}       &                                                             & \textbf{Performance across top 4 NFR classes (249 Requirements)}                &              &               & \textbf{\textbf{Weighted Average~}} &               &                \\
                                  &                                                             & \textbf{P}                                  & \textbf{R}   & \textbf{F1}   & \textbf{wP}                 & \textbf{wR}   & \textbf{wF1}   \\
Llama                             & Is-about assertion                                         & 0.3588                                      & 0.4000       & 0.3164        & 0.3731                      & 0.4257        & 0.3348         \\
All-Mini                          & Expert curated                                              & 0.6609                                      & 0.6455       & 0.6250        & 0.6648                      & 0.6426        & 0.6238         
\end{tblr}
\end{table}

\begin{table}
\centering
\caption{Comparison of Top-Performing Generative and Non-Generative LLMs on Multi-Class Classification Tasks for the Top 4 Classes.}
\label{tab:Top4-results}
\scriptsize 

\begin{tblr}{
  width = \linewidth,
  colspec = {Q[63]Q[121]Q[62]Q[62]Q[62]Q[62]Q[62]Q[62]Q[62]Q[62]Q[62]Q[62]Q[62]Q[62]},
  cell{1}{1} = {c,t},
  cell{1}{2} = {t},
  cell{1}{3} = {c=12}{0.744\linewidth,c},
  cell{2}{3} = {c=3}{0.186\linewidth,c},
  cell{2}{6} = {c=3}{0.186\linewidth,c},
  cell{2}{9} = {c=3}{0.186\linewidth,c},
  cell{2}{12} = {c=3}{0.186\linewidth,c},
  cell{3}{3} = {c},
  cell{3}{4} = {c},
  cell{3}{5} = {c},
  cell{3}{6} = {c},
  cell{3}{7} = {c},
  cell{3}{8} = {c},
  cell{3}{9} = {c},
  cell{3}{10} = {c},
  cell{3}{11} = {c},
  cell{3}{12} = {c},
  cell{3}{13} = {c},
  cell{3}{14} = {c},
  hline{1,6} = {-}{0.08em},
  hline{2} = {-}{},
  hline{3-4} = {3-14}{},
}

{\textbf{Model}} & {\textbf{Prompt/Label}} & \textbf{Performance on The Top 4 Individual NFR Classes} &              &               &                          &              &               &                             &              &               &                             &              &               \\
                 &                                       & \textbf{ Usability (67 Requirements) }            &              &               & \textbf{Security (66 Requirements) } &              &               & \textbf{ Operational (62 Requirements) } &              &               & \textbf{ Performance (54 Requirements) } &              &               \\
                 &                                       & \textbf{ P }                         & \textbf{ R } & \textbf{ F1 } & \textbf{ P }             & \textbf{ R } & \textbf{ F1 } & \textbf{ P }                & \textbf{ R } & \textbf{ F1 } & \textbf{ P }                & \textbf{ R } & \textbf{ F1 } \\
Llama            & Is-about assertion                  & 0.3557                               & 0.7910       & 0.4907        & 0.5341                   & 0.7121       & 0.6104        & 0.5455                      & 0.0968       & 0.1644        & 0.0000                      & 0.0000       & 0.0000        \\
All-Mini         & Expert curated                        & 0.7742                               & 0.3582       & 0.4898        & 0.6327                   & 0.9394       & 0.7561        & 0.6424                      & 0.5806       & 0.6102        & 0.5938                      & 0.7037       & 0.6441        
\end{tblr}
\end{table}

\begin{itemize}
    \item \textbf{Task NFR:} The goal of this task is to classify each requirement into one of the 10 NFR classes. As shown in Table \ref{tab:multi-best-results}, All-Mini emerges as the top-performing model, delivering consistent results in both precision and recall. In contrast, Llama performs poorly, struggling with both precision and recall.
    
    \item \textbf{Task NFR-Top4:} The goal of this task is to classify each requirement into one of the top 4 classes. As shown in Table \ref{tab:multi-best-results}, All-Mini again stands out as the top-performing model, delivering consistent results in both precision and recall. In contrast, Llama performs poorly, struggling with both precision and recall. A deeper analysis of these models' performance on individual classes, as seen in Table \ref{tab:Top4-results}, reveals that \textbf{Llama} fails to classify any requirements into the Performance class, achieving 0.00\% on all metrics. In contrast, All-Mini excels in this class, particularly in recall. Furthermore, All-Mini demonstrates strong performance in the Usability class, especially with precision (0.7742\%), and achieves impressive recall (0.9394\%) on the Security class. These results clearly demonstrate that All-Mini is the top-performing model for this task.

\end{itemize}

In summary, the evaluation demonstrates that \textbf{All-Mini} is the clear winner in both multi-class classification tasks, delivering strong performance that \textbf{Llama} cannot match.

\subsection{Statistical Signifcance Analysis}

\begin{table}
\centering
\caption{Wilcoxon test results comparing the best-performing models (BEST LLMs) from the embedding-based and inference-based approaches, for each task, assessing statistical significance. Cells marked by (*) refer to a rejection of the null hypothesis when $p$-value $< 0.05$.}
\label{tab:stat-wilcoxon}
\scriptsize 
\begin{tblr}{
  row{1} = {c},
  row{2} = {c},
  row{6} = {c},
  row{7} = {c},
  cell{1}{1} = {c=4}{},
  cell{3}{1} = {r=3}{c},
  cell{3}{4} = {c},
  cell{4}{4} = {c},
  cell{5}{4} = {c},
  cell{6}{1} = {c=4}{},
  cell{8}{1} = {r=2}{c},
  cell{8}{4} = {c},
  cell{9}{4} = {c},
  hline{1-3,6-8,10} = {-}{},
}
\textbf{\textbf{A) Binary Classification Results}}                        &               &                    &                    \\
\textbf{Measure~}                                                         & \textbf{Task} & \textbf{Best LLMs} & \textbf{$p$-value} \\
$wF1$                                                                     & Functional    & Llama vs. SBERT    & 0.01600 *          \\
                                                                          & Quality       & Gemma vs. All-Mini & 0.00400 *          \\
                                                                          & Security      & Gemma vs. All-Mini & 0.03300 *          \\
\textbf{\textbf{B) Multi-class Classification~\textbf{\textbf{Results}}}} &               &                    &                    \\
\textbf{Measure}                                                          & \textbf{Task} & \textbf{Best LLMs} & \textbf{$p$-value} \\
$wF1$                                                                     & NFR           & Llama vs. All-Mini & 0.01000 *          \\
                                                                          & NFR-Top4      & Llama vs. All-Mini & 0.00000 *          
\end{tblr}
\end{table}

The results from the Wilcoxon Signed-Rank Test are presented in Table \ref{tab:stat-wilcoxon}, which shows that the \textbf{$p$-value is $\leq 0.05$} for all binary and multi-class classification tasks. This indicates that the observed differences in performance between the best generative LLM and the best non-generative LLM are \textit{statistically significant}. While the Wilcoxon Test does not indicate which model performs better, it confirms that the performance of the non-generative and generative LLMs differs significantly across all five requirements classification tasks.

\subsection{Addressing Research Question RQ2}

Based on our experimental results and statistical analysis, we can now address RQ2:

\begin{enumerate}
    
\item \textbf{No Clear Winner in Binary Classification Tasks:}

\begin{itemize}
    \item Among the generative models (Llama and Gemma) and non-generative models (SBERT and All-Mini), there is no clear overall winner.
    \item Llama outperforms other models on Task Functional and Gemma excels in Task Quality.
    \item All-Mini outperforms Gemma on Task Security.
    \item The prompt pattern also impacts performance, with different patterns working better for different models.
\end{itemize}

\item \textbf{Clear Winner in Multi-Class Classification Tasks:}

\begin{itemize}
    \item All-Mini outperforms Llama in both multi-class classification tasks, showing consistent strong performance.
    \item Llama struggles significantly, especially on Task NFR and Task NFR-Top4, with particularly poor performance on the Performance class in Task NFR-Top4.
    \item All-Mini demonstrates strong precision and recall, particularly excelling in the Usability and Security classes.
\end{itemize}

\item \textbf{Generative vs Non-Generative Models:}

\begin{itemize}

    \item \textbf{Generative models:} Llama outperforms other models on Task Functional, while Gemma performs best on Task Quality. However, neither generative nor non-generative models excels across all binary classification tasks.
    \item\textbf{Non-generative models:} All-Mini demonstrates strong performance in both multi-class classification tasks and excels in one binary classification task, making it the top-performing model overall across both binary and multi-class classification tasks.
\end{itemize}

\end{enumerate}

The key findings for RQ2 are summarised below.

\begin{tcolorbox}[colback=green!5!white,colframe=green!75!black,title=Key Findings for RQ2]\label{RQ2Key}
\small
  \begin{itemize}
    \item In the case of binary classification, no single model emerges as the definitive winner, as each model excels in different tasks and contexts.
    \item For multi-class classification, All-Mini stands out as the top performer, demonstrating strong and consistent results across tasks. In contrast, Llama struggles to match its performance, particularly in handling specific class predictions.
    \item Overall, the non-generative model, All-Mini, outperforms both generative models in requirements classification. 
\end{itemize}

\end{tcolorbox}

\section{Validity Threats}
\label{sec:valid}

In our experimental studies, several potential threats to validity could undermine the accuracy and reliability of our conclusions. These threats may affect \textbf{internal validity} (the ability to draw causal conclusions within the study), \textbf{external validity} (the extent to which results can be generalized to other settings or populations), and \textbf{construct validity} (the accuracy with which the study measures the intended variables). This section outlines the strategies we have employed to mitigate these threats and enhance the validity of our findings across all three dimensions.

\subsection{Internal Validity}

To enhance the internal validity of our studies, we implemented several key strategies designed to minimise potential confounds and biases. Internal validity is essential as it ensures that observed effects can be confidently attributed to the experimental manipulation, rather than to extraneous factors. The strategies we employed include:

\begin{itemize} 
\item \textbf{Standardized procedures:} We adhered to established experimental protocols to ensure that all experiments were conducted under identical conditions. By standardising procedures, we minimised variability across experiments and ensured that the results were due to the experimental manipulation, rather than inconsistencies in how the study was conducted. 

\item \textbf{Dataset selection:} We carefully selected three datasets that are representative of a broad spectrum of requirements classification tasks, aiming to minimise dataset bias. While the quality of the PROMISE dataset used in our study may impact the performance of the LLMs, it is widely regarded as a \textit{de facto} benchmark within the RE community. Using these datasets allows other researchers to compare their results with ours, enhancing the external validity of our findings.

\item \textbf{Model selection:} To reduce model bias, we carefully selected three diverse LLMs for our studies, each with different attention mechanisms and training data. This approach ensures that our results are not overly influenced by the characteristics of any single model, providing a more comprehensive assessment of LLM performance.

\item \textbf{Replication and transparency:} To strengthen the robustness of our findings, we made all data, code, and experimental protocols publicly available on GitHub. This transparency facilitates the replication of our study by other researchers and strengthens the credibility of our conclusions.

\item \textbf{Minimizing confounding variables:} Confounding variables—factors that may influence both the independent variable (e.g., LLMs or experimental conditions) and the dependent variable (e.g., classification accuracy)—can lead to incorrect conclusions about causality. To minimise the impact of such variables, we ensured that all LLMs were evaluated on the same datasets and measured using consistent evaluation metrics.

\end{itemize}

\subsection{External Validity}

While we believe that our findings are generalisable to similar contexts where comparable tasks are performed, we acknowledge two key external validity threats that may limit the broader applicability of our conclusions:

\begin{itemize}

\item \textbf{Technological obsolescence:} The LLMs used in our study may soon become outdated as new, more advanced models are developed. The rapid pace of advancements in artificial intelligence means that the models tested in our study may not be representative of the state-of-the-art models in the future. This potential for technological obsolescence poses a challenge to the long-term relevance of our findings, as newer models could outperform or behave differently from the models we used. However, the insights from our study—specifically regarding the general performance of LLMs on requirements classification tasks—will still serve as a valuable baseline for comparison with future advancements.

\item \textbf{Dataset limitations:} The quality and characteristics of the datasets used in our study, particularly the PROMISE dataset, may influence the performance of the LLMs. While the PROMISE dataset is widely recognised as a \textit{de facto} benchmark in RE community, it is not without limitations. The dataset may not fully capture the diversity or complexity of real-world requirements, and its inherent biases could affect the generalisability of our findings to other datasets or contexts. Nevertheless, the widespread use of the PROMISE dataset ensures that our results remain comparable to those of other studies in the RE community. Therefore, our findings can serve as a reliable point of reference for future research that employs different datasets.

\end{itemize}

By ensuring the transparency of our study and making our data and code publicly available, we aim to contribute to a broader understanding of the role of LLMs in requirements classification. At the same time, we recognise that external validity is an ongoing concern and encourage future research to replicate and extend our work across varied datasets and evolving models.

\subsection{Construct Validity}

To address potential threats to construct validity, we implemented several strategies to ensure that our study accurately measured the intended variables and that the observed effects truly reflected the constructs being tested. The strategies we employed to strengthen construct validity include:

\begin{itemize} 

\item \textbf{Robust statistical testing:} A central component of our analysis was the use of the Friedman test, a non-parametric statistical method that allowed us to rigorously analyse the effects of key experimental constructs—namely, LLMs, prompt patterns, datasets, and tasks—without assuming normality in the data. This was especially important given the non-uniform performance scores observed across different experimental conditions. The Friedman test enabled us to account for inherent variability in the data and draw valid conclusions about the relationships between these experimental variables, ensuring that the results were not skewed by outliers or distributional assumptions. We also applied the Wilcoxon Signed-Rank Test to confirm that the differences in the results obtained from the generative LLMs and the non-generative LLMs are statistically significant across both binary and multi-class classification tasks.

\item \textbf{Clarity and consistency in construct operationalisation:} To further safeguard construct validity, we ensured that the key experimental constructs—such as LLMs, datasets, and taskas wass—were clearly defined and consistently measured across all conditions. We carefully selected well-established datasets, such as the PROMISE dataset, which is widely regarded as a benchmark in RE community. This ensured that the datasets used were not only relevant but also reliable for comparison across studies. Additionally, we standardized task descriptions and ensured they were uniformly understood across all experimental trials, minimizing any potential ambiguity in how the constructs were interpreted and measured.

\end{itemize}

By combining rigorous statistical methods with a clear, consistent operationalisation of the key constructs, we strengthened the construct validity of our study. This approach enabled us to confidently draw conclusions that accurately reflect the effects of the experimental constructs, rather than being influenced by external factors or measurement inconsistencies.

For full details on the experimental setup, please refer to the online supplementary materials\footnote{\url{https://github.com/waadalhoshan/LLM4RC}}.

\section{Conclusion}
\label{sec:concl}

In this study, we conducted a comprehensive and systematic evaluation of generative LLMs for requirements classification tasks, involving over 400 experiments with 3 generative models (Bloom, Gemma, and Llama), 5 classification task groups, and 3 datasets. These tasks included 3 binary classification tasks (Functional, Quality, and Security) and 2 multi-class classification tasks (Task NFRs and Task NFR-Top4). Additionally, we applied 5 variations to the datasets, including modifications at both the requirement statement and label representation levels.

Our findings highlight several key insights:

\begin{itemize} 

\item The choice of LLMs significantly affects performance. Bloom's self-attention architecture favours high precision but may compromise recall, while Gemma excels in binary classification tasks, particularly in recall, but struggles with multi-class classification. Llama's optimised transformer architecture delivers strong multi-class classification performance and balanced results in binary tasks. 

\item Prompt formulation plays a critical role in classification outcomes. Assertion-based prompts are highly effective across both binary and multi-class tasks due to their simplicity and clarity. Definition-based prompts, on the other hand, are particularly useful for distinguishing closely related labels in multi-class scenarios. 

\item Generative LLMs are robust to dataset variations, maintaining consistent performance despite changes in requirement text and label formatting. This demonstrates the models' ability to focus on semantic content rather than specific textual presentation. 

\end{itemize}

When comparing generative LLMs to non-generative models (All-Mini and SBERT), we found that no single model consistently outperformed the others in binary classification tasks. However, in multi-class classification, All-Mini emerged as the top performer, while Llama struggled to match its results. Overall, non-generative models, particularly All-Mini, outperformed generative models in requirements classification, supporting our hypothesis regarding the effectiveness of non-generative models for this task.

Future research should explore cross-domain generalisation, investigate the impact of model scaling on diverse text classification tasks, and examine the application of these models in classifying app reviews. These directions will further enhance our understanding of generative LLMs' capabilities and limitations, improving their practical applicability across various domains. In conclusion, this work lays the groundwork for further exploration of generative LLMs in the field of requirements engineering.


\begin{acks}
This work was supported and funded by the Deanship of Scientific Research at Imam Mohammad Ibn Saud Islamic University (IMSIU) (grant number IMSIU-DDRSP2504).
\end{acks}

\bibliographystyle{ACM-Reference-Format}





\end{document}